\documentclass[a4paper]{cas-sc}

\usepackage[authoryear]{natbib}
\ExplSyntaxOn

\ExplSyntaxOff

\usepackage{amsmath,amssymb,amsthm}
\usepackage{algorithm,algorithmic}
\usepackage{wrapfig}
\usepackage{graphicx}
\newtheorem{definition}{Definition}
\newtheorem{assumption}{Assumption}
\newtheorem{proposition}{Proposition}
\newtheorem{corollary}{Corollary}
\newcommand{\ourmethod}{Doctor}
\def\tsc#1{\csdef{#1}{\textsc{\lowercase{#1}}\xspace}}
\tsc{WGM}
\tsc{QE}
\begin{document}
\let\WriteBookmarks\relax
\def\floatpagepagefraction{1}
\def\textpagefraction{.001}

% Short titleNeural
\shorttitle{}    

\shortauthors{}  

\title [mode = title]{Hybrid Sequence Modeling and Reinforced Verification for Controllable Target-Conditioned Decision Making}  
\shortauthors{Pei et al.}
\author[1]{Yue Pei}
% \credit{Conceptualization, Methodology, Software, Validation, Formal analysis, Investigation, Writing -- original draft}

\author[2]{Hongming Zhang}
% \credit{Methodology, Software, Validation, Writing -- review \& editing}

\author[3]{Chao Gao}
% \credit{Methodology, Writing -- review \& editing}

\author[2]{Martin M{\"u}ller}
% \credit{Supervision, Writing -- review \& editing}

\author[4]{Yingying Zhang}
\cormark[1]
% \ead{zhangyingying@buaa.edu.cn}
% \credit{Supervision, Writing -- review \& editing}

\author[5]{Mengxiao Zhu}
% \credit{Supervision, Writing -- review \& editing}

\author[6]{Ziliang Chen}
% \credit{Supervision, Writing -- review \& editing}

\author[4]{Hao Sheng}
% \credit{Supervision, Writing -- review \& editing}

\author[6]{Liang Lin}
% \credit{Supervision, Writing -- review \& editing}

\author[4]{Haogang Zhu}
\cormark[1]
% \ead{haogangzhu@buaa.edu.cn}
% \credit{Funding acquisition, Project administration, Writing -- review \& editing}

% Affiliations
\affiliation[1]{
    organization={School of Artificial Intelligence, Beihang University},
    city={Beijing},
    country={China}
}

\affiliation[2]{
    organization={Department of Computing Science and Amii, University of Alberta},
    city={Edmonton},
    country={Canada}
}

\affiliation[3]{
    organization={Edmonton Research Center, Huawei Canada},
    city={Edmonton},
    country={Canada}
}

\affiliation[4]{
    organization={Hangzhou International Innovation Institute, Beihang University},
    city={Hangzhou},
    country={China}
}

\affiliation[5]{
    organization={School of Artificial Intelligence and Computer Science, North China University of Technology},
    city={Beijing},
    country={China}
}

\affiliation[6]{
    organization={Research Institute of Multiple Agents and Embodied Intelligence, Peng Cheng Laboratory},
    city={Shenzhen},
    country={China}
}

\cortext[1]{Corresponding author}

\begin{abstract}
Target-conditioned sequence models provide a simple interface for controllable offline decision making, but the requested target return can be an unreliable control signal, especially when the target return lies in underrepresented regions of the dataset. This paper proposes \ourmethod, a hybrid sequence modeling and reinforced verification framework for controllable target-conditioned offline decision making. \ourmethod\ trains a shared masked trajectory Transformer with two complementary objectives: masked trajectory reconstruction for candidate generation and in-sample value learning for action-value verification. At inference time, the model samples multiple nearby target returns, generates candidate actions in parallel, and selects the action whose verified value is closest to the requested target return. We analyze this verifier-guided selection rule and show that its value-level alignment error is bounded by candidate-value coverage around the target return and verifier accuracy. Experiments on D4RL and EpiCare show that \ourmethod\ improves target-return alignment under reduced high-return coverage, remains competitive on standard offline return-maximization benchmarks, and enables a single policy to modulate between conservative and aggressive operating points in a simulated clinical decision-making task. These results suggest that reinforced verification can improve the controllability of target-conditioned policies.
\end{abstract}


% \begin{highlights}
% \item 
% \item 
% \item 
% \end{highlights}

% Keywords
% \begin{keywords}
%  \sep \sep \sep
% \end{keywords}


\maketitle

\section{Introduction}
\label{sec:introduction}

% Sequential decision models have become increasingly expressive, with recent
% sequence-modeling and transformer-based approaches showing strong ability to learn
% policies from trajectory data, capture long-horizon dependencies, and optimize
% cumulative returns in complex decision-making tasks \citep{matsuo2022deep,liu2024adaptive,zhang2025diffusion}. Much of this progress has
% been driven by methods that focus on a prescribed optimization objective, typically
% extracting a policy that achieves high return under the available data and task
% definition. In many decision-making systems, the desired policy is controllable and can
% operate at different performance levels, risk preferences, or intervention
% strengths. Such controllability is essential for trustworthy deployment: a model
% should not only generate plausible actions but also make its realized
% consequences align with the target requested by the user.

Sequential decision models have become increasingly expressive, with recent neural approaches improving value estimation, representation learning, and Transformer-based policy modeling for complex reinforcement learning tasks \citep{zhang2024unified,wang2024romat,feng2025timar}. These advances have strengthened the ability of learned policies to exploit trajectory data, capture long-horizon dependencies, and optimize cumulative returns in complex decision-making problems. Much of this progress has been driven by methods that focus on a prescribed optimization objective, typically extracting a policy that achieves high return under the available data and task definition. In many decision-making systems, however, the desired policy should be controllable and able to operate at different performance levels, risk preferences, or intervention strengths. Such controllability is essential for trustworthy deployment: a model should not only generate plausible actions but also make its realized consequences align with the target requested by the user.

This requirement appears across a wide range of sequential decision problems. In
clinical decision-making, a treatment policy may need to trade therapeutic benefit
against adverse-event risk according to patient condition and clinical preference
\citep{goetz2018personalized,hargrave2024epicare}. In education, an intelligent
tutor may need to adjust content difficulty to the learner's current ability
\citep{singla2021reinforcement,ALAWWAD2025111332}. In game AI and robotic control,
one may prefer a single policy that can express multiple skill levels rather than
one uniformly strongest agent \citep{jeon2023raidenv}. These examples require
\emph{target-conditioned control}: given a desired return or operating point, the
policy should produce behavior whose realized outcome follows that target. We
refer to this requirement as \emph{target alignment}.

A common instantiation of target-conditioned control is to use a desired return as the target. Return-conditioned sequence modeling provides a natural interface for this instantiation: reinforcement learning via supervised learning methods formulate offline reinforcement learning as conditional sequence modeling, where a policy predicts actions from trajectory context and a desired target return \citep{chen2021decision,janner2021offline,emmons2021rvs}. By changing this prompt, a single trained model can in principle be queried for different performance levels. However, the target return is not automatically a calibrated control signal. In offline RL, the model is learned from a fixed dataset without additional interaction with the environment \citep{levine2020offline,fu2020d4rl}, and the dataset often provides uneven coverage across return levels. As a result, a target-conditioned policy may learn correlations between target-return tokens and observed actions, but still generate actions whose expected or realized returns deviate from the requested target return, especially when the target return lies in an underrepresented return region or near the boundary of the observed return range. Such target-alignment failures have been observed in recent studies of return-conditioned offline RL \citep{tanaka2024return,brandfonbrener2022does}. Value-based offline RL offers a complementary strength: temporal difference learning evaluates actions through their long-term consequences and can stitch useful behaviors across different trajectories. This motivates our reinforced verification mechanism, which couples target-conditioned generation with a value verifier and uses verifier-guided selection to choose the candidate whose predicted value is closest to the requested target return before execution.

In this paper, we propose \ourmethod, a hybrid sequence modeling and reinforced
verification framework for controllable target-conditioned decision making.
\ourmethod\ separates decision-making into two complementary pathways: a
generative pathway and an evaluative pathway. The generative pathway is a
bidirectional transformer trained with masked trajectory reconstruction, which
proposes candidate actions under sampled target returns around the requested
target return. The evaluative pathway is an in-sample temporal-difference value head,
which serves as a verifier of the predicted consequence of each candidate action.
At inference time, instead of directly executing the action produced from a single
target-return prompt, \ourmethod\ generates multiple target-conditioned candidate actions
in parallel and executes the candidate whose verifier value is closest to the
requested target return. This reinforced verification turns a target return from a weak
conditioning signal into a value-checked control interface. We further provide a
decision-level theoretical analysis showing that the alignment error of this reinforced verification is controlled by two factors: how well the generated
candidate set covers actions near the requested target value and how accurately
the verifier estimates candidate values. 
% The analysis also explains why increasing
% the number of generated candidates can improve target alignment until verifier
% error becomes the limiting factor.

We evaluate \ourmethod\ on standard offline RL benchmarks from D4RL and on the EpiCare benchmark for dynamic treatment regimes \citep{fu2020d4rl,hargrave2024epicare}. The experiments examine both target-return alignment and downstream control. On locomotion tasks with deliberately reduced high-return coverage, \ourmethod\ achieves realized returns that are closer to the requested targets than return-conditioned sequence modeling baselines. Ablations show that both masked trajectory modeling and the reinforced verification contribute to lower alignment error, and increasing the candidate budget improves alignment in a manner qualitatively consistent with the coverage analysis. Beyond alignment evaluation, \ourmethod\ also achieves competitive return-maximization performance on locomotion, Maze2D, and Adroit tasks. On EpiCare, varying the target return modulates the operating point of a single trained policy, producing a meaningful trade-off between clinical return and adverse-event risk in the simulated benchmark.

Our contributions are as follows: (1) we formulate controllable target-conditioned
offline decision making as a target-alignment problem, where the realized policy
behavior should match a user-specified target;
(2) we propose \ourmethod, a hybrid framework that couples masked trajectory
reconstruction with in-sample temporal-difference value learning for reinforced
verification; (3) we introduce an inference-time reinforced verification mechanism that
generates multiple target-conditioned candidates and selects the one whose
verifier value is closest to the requested target; (4) we provide a conditional
decision-level analysis showing that alignment depends on candidate-value coverage
and verifier accuracy; and (5) we empirically demonstrate episodic return
alignment, competitive return-maximization performance on D4RL, and controllable
policy modulation on the simulated EpiCare benchmark.

% #####################################################################
\section{Problem Setup}
\label{sec:prelim}
% #####################################################################

\subsection{Offline reinforcement learning and return-conditioned modeling}
\label{sec:prelim-offline-rvs}

We consider a finite-horizon Markov Decision Process (MDP)
$\mathcal M=(\mathcal S,\mathcal A,\mathcal R,P,\gamma)$, where
$\mathcal S$ is the state space, $\mathcal A$ is the action space,
$\mathcal R(s,a)$ is a bounded reward function, $P(s'\mid s,a)$ is the transition
kernel, and $\gamma\in[0,1]$ is the discount factor. A trajectory is
$\tau=(s_0,a_0,r_0,\ldots,s_T,a_T,r_T)$, and the dataset return-to-go at timestep
$t$ is $R_t^{\mathcal D}=\sum_{i=t}^{T}\gamma^{i-t}r_i .$
In offline RL, the agent only has access to a static dataset
$\mathcal D=\{\tau_j\}$ collected by unknown behavior policies. No additional
interaction is available during offline training.
Return-conditioned sequence models, such as Decision Transformer, train a policy
by supervised learning on trajectory sequences and condition action prediction on
a desired target return. For a context length $K$, the training sequence around
$t$ is
\begin{equation}
\tau_t^{K}=\bigl(R_{t-K+1}^{\mathcal D},s_{t-K+1},a_{t-K+1},\ldots,
R_t^{\mathcal D},s_t,a_t\bigr).
\label{eq:prelim-sequence}
\end{equation}
% At inference time, the dataset return token is replaced by a requested target
% return. To avoid notational ambiguity, we use $R_t^{\mathcal D}$ for the return-to-go
% stored in the offline dataset and $g_t$ for the desired target return specified at
% inference.
At inference time, the dataset return token is replaced by a requested target
return. To avoid notational ambiguity, we use $R_t^{\mathcal D}$ for the return-to-go
stored in the offline dataset and $g_t$ for the desired target return specified at
inference. 
% Throughout the method and analysis, the same discount factor $\gamma$
% is used for the dataset return-to-go $R_t^{\mathcal D}$, the value objective
% in Eq.~\eqref{eq:vloss}, and the inference-time target update in Eq.~\eqref{eq:rtg-update}; the verifier target and the value estimate therefore follow the same return-to-go convention.
Because our transformer uses a finite history, we write
\begin{equation}
h_t=(s_{t-K+1},a_{t-K+1},r_{t-K+1},\ldots,s_t).
\label{eq:context}
\end{equation}
% for the decision context available before choosing $a_t$. In a fully observed
% Markov setting, $h_t$ can be read simply as $s_t$. 
We use this context notation in
the analysis below.

\subsection{Value-level target alignment}
\label{sec:prelim-alignment}

We formalize target alignment at the level of action values. Given a decision
context $h$ and a requested target return $g$, an aligned target-conditioned
policy should choose an action whose expected long-term return is close to $g$.
This requires specifying both the value against which alignment is measured and
the set of target values that are attainable at the current context. Let $Q^*(h,a)$ denote the ideal action-value used in our analysis. It represents
the expected return obtained by taking action $a$ at context $h$ and then
following the best continuation within the action region on which value
evaluation is reliable. Equivalently, $Q^*$ can be characterized by the
in-sample Bellman relation
\begin{equation}
Q^*(h_t,a_t)
=
\mathbb E\!\left[
r_t+\gamma
\max_{a'\in\mathcal A_{\mathrm{val}}(h_{t+1})}
Q^*(h_{t+1},a')
\,\middle|\, h_t,a_t
\right],
\label{eq:insample}
\end{equation}
where $\mathcal A_{\mathrm{val}}(h)\subseteq\mathcal A$ denotes the reliably
evaluable action region. 
% We treat \eqref{eq:insample} as the defining relation of the oracle value used in our analysis. In a fully observed setting the context reduces to the current state, $h_t\equiv s_t$, and \eqref{eq:insample} is the usual in-sample Bellman relation. In
% the partially observed case, $h_t$ is a finite history and \eqref{eq:insample} is understood as an idealized characterization on the context process rather than an assertion that $(h_t,a_t)$ is exactly Markov; the guarantees below only use $Q^*$ through its values on the finite candidate set evaluated at inference.
% This set is only a modelling device for defining the
% oracle value $Q^*$; \ourmethod\ does not require knowing it explicitly.
For a fixed context $h$, not every scalar target value need be attainable by an
action. We therefore define the attainable value image as
\begin{equation}
\mathcal V(h)
=
\overline{\{\,Q^*(h,a):a\in\mathcal A_{\mathrm{val}}(h)\,\}},
\label{eq:value-image}
\end{equation}
where $\overline{\cdot}$ denotes closure. Thus, a target $g$ is feasible at
$h$ if it can be matched or arbitrarily approximated by some attainable action
value, i.e.,
\begin{equation}
\inf_{v\in\mathcal V(h)} |v-g|=0 .
\end{equation}
This value-image formulation avoids assuming that all values in an interval are
attainable, which may fail for discrete or disconnected action spaces.
\begin{definition}[Value-level alignment error]
\label{def:alignment-error}
For a decision context $h$, target return $g$, and executed action $a$, the
value-level target alignment error is
\begin{equation}
\mathcal E(h,g;a)=\bigl|g-Q^*(h,a)\bigr|.
\label{eq:value-level-error}
\end{equation}
\end{definition}
Definition~\ref{def:alignment-error} measures whether the selected action is
value-consistent with the requested target at a single decision point. This is
different from measuring the realized return of one stochastic rollout: the
latter may vary due to transition noise, whereas $\mathcal E(h,g;a)$ isolates
the value mismatch caused by action selection. The verifier-guided selection in
Section~\ref{sec:prelim-rule} is designed to reduce this quantity by selecting,
among generated candidates, the action whose estimated value is closest to $g$.
The analysis in Section~\ref{sec:theory} then bounds this error in terms of
candidate-value coverage and verifier accuracy.

\subsection{Target-conditioned generation and reinforced verification}
\label{sec:prelim-rule}

\ourmethod\ combines two components that play different roles in target alignment.
The generative pathway $\pi_{\mathrm{gen}}(a\mid h,g)$ proposes actions under a
target return $g$, while the evaluative pathway $Q_\phi(h,a)$ verifies the
predicted value of each proposed action. Thus, the target return is used to
generate candidates, and the verifier is used to decide which candidate best
matches the requested target.

\begin{definition}[Verifier-guided selection]
\label{def:verifier-rule}
Given context $h$, target $g$, sampling budget $N$, and bandwidth $\delta>0$, let
$B_\delta(g)=\{g':|g'-g|\le\delta\}$. \ourmethod\ samples nearby prompts and
generates candidate actions,
% \begin{align}
% g_i&\sim\mathrm{Unif}\bigl(B_\delta(g)\bigr),
% &
% a_i&\sim\pi_{\mathrm{gen}}(\cdot\mid h,g_i),
% \qquad i=1,\ldots,N.
% \end{align}
\begin{equation}
g_i \sim \mathrm{Unif}\bigl(B_\delta(g)\bigr),
\qquad
a_i \sim \pi_{\mathrm{gen}}(\cdot\mid h,g_i),
\qquad
i=1,\ldots,N.
\label{eq:sample-generate}
\end{equation}
It then executes the candidate whose verifier value is closest to the requested
target,
\begin{equation}
i^*\in\arg\min_{i\in[N]}\bigl|Q_\phi(h,a_i)-g\bigr|,
\qquad
\pi_{\mathrm{ver}}(h,g)=a_{i^*}.
\label{eq:verifier-rule}
\end{equation}
\end{definition}
Definition~\ref{def:verifier-rule} formalizes the reinforced verification of
\ourmethod. The generated candidate set
controls whether actions near the requested target value are available, and the
verifier controls whether such actions can be selected reliably. These two factors
correspond to candidate-value coverage and verifier accuracy in the theoretical
analysis.

% #####################################################################

\section{Method}\label{sec:method}
 
Section~\ref{sec:prelim} introduces the two components needed for target alignment: a
target-conditioned generator $\pi_{\mathrm{gen}}(a\mid h,g)$ that proposes actions
under a requested target return, and a verifier $Q_\phi(h,a)$ that estimates the
value of each proposed action. \ourmethod\ instantiates these components with a
shared masked trajectory transformer. The generator produces candidate actions
conditioned on nearby target returns, while the verifier evaluates these candidates with the learned value estimator. At inference time, the final action is chosen
by comparing the verifier values with the requested target.

\begin{figure}
	\centering
	\includegraphics[width=0.9\textwidth]{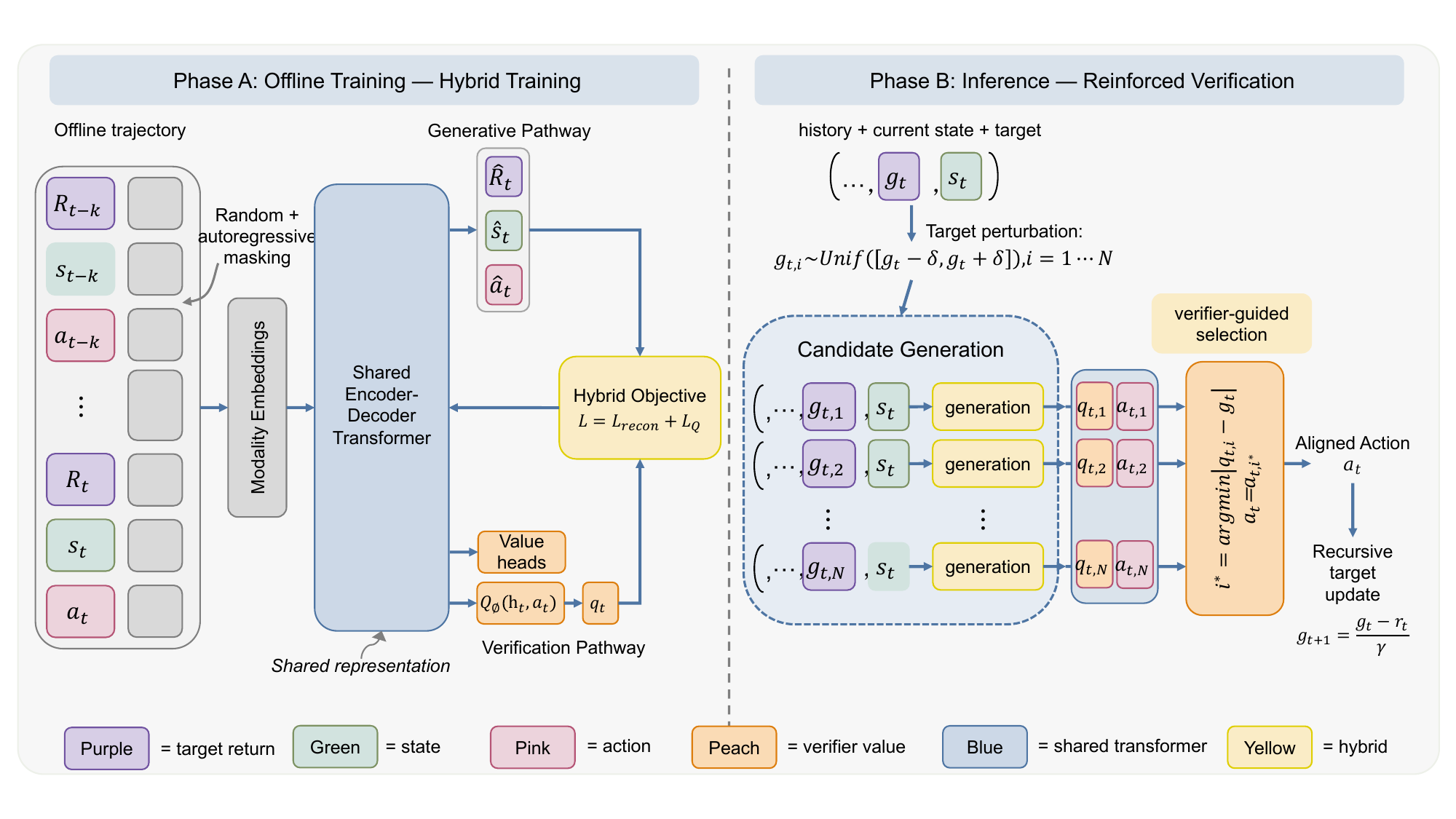}
	\caption{Overview of \ourmethod. Left: during offline training, returns, states, and actions are embedded and processed by a shared encoder-decoder Transformer under partial trajectory masking. Reconstruction heads recover masked trajectory tokens, while value heads learn to verify action values through an in-sample TD objective. Right: during inference, \ourmethod\ samples nearby target returns, generates multiple candidate actions in parallel, scores them with the verifier, and executes the action whose predicted value is closest to the requested target.}
	\label{fig:overview}
\end{figure}

This design separates two roles that are entangled in a standard
return-conditioned policy. The target return guides candidate generation, but the
executed action is not determined by the prompt alone. Instead, each generated
candidate is evaluated by the verifier before execution. This mechanism makes the target return a control signal calibrated through reinforced verification rather than only a conditioning token. The rest of this section describes the shared model
architecture, the hybrid training objective, and the inference-time
reinforced verification. Figure~\ref{fig:overview} summarizes the overall architecture of \ourmethod, including the offline hybrid training pipeline and the reinforced verification procedure.

% ---------------------------------------------------------------------
\subsection{Model architecture}\label{sec:method-arch}
% ---------------------------------------------------------------------
 
\ourmethod\ builds both pathways on one encoder-decoder transformer, which serves as a
shared trajectory representation extractor. Each return, state, and action token
is first embedded by a modality-specific layer. During training, the length-$K$
trajectory sequence is corrupted with a mixture of random and autoregressive
masks,
\begin{equation}
M(\tau_t^{K})=\bigl(R_{t-K+1}^{\mathcal D},\;\rule[-0.5ex]{0.2cm}{0.5pt}\;,
a_{t-K+1},\;\rule[-0.5ex]{0.2cm}{0.5pt}\;,s_{t-K+2},\ldots,
R_t^{\mathcal D},s_t,\;\rule[-0.5ex]{0.2cm}{0.5pt}\;\bigr),
\label{eq:mask}
\end{equation}
where $\rule[-0.5ex]{0.2cm}{0.5pt}$ denotes a masked element. The encoder
$E_\theta$ and decoder $D_\theta$ map the masked sequence to a latent
trajectory representation
\begin{equation}
z_t=D_\theta\bigl(E_\theta(M(\tau_t^{K}))\bigr),
\end{equation}
from which lightweight linear or MLP heads reconstruct the return, state, and
action tokens.
 
\paragraph{Generative pathway.}
The action head implements the target-conditioned generator
$\pi_{\mathrm{gen}}(a\mid h,g)$. During training, it learns to reconstruct masked
actions from the surrounding trajectory context. During inference, the current
return token $R_t^{\mathcal D}$ is replaced by a target return $g$, and the
current action slot is masked. The action head then outputs a conditional action
distribution under the finite decision context $h_t$. In this way, the same
trained model can be queried with different target returns to produce different
candidate actions.
 
\paragraph{Evaluative pathway.}
A per-timestep value head with parameters $\phi$ is attached to the same shared
representation and defines the verifier $Q_\phi(h_t,a_t)$. The sampled target-return prompt determines which candidate action is generated, but the verifier evaluates the
resulting context--action pair. This distinction is important because the
selection step should compare the predicted value of each candidate with the
requested target, rather than simply trusting the prompt that produced the
candidate. Sharing the transformer representation allows the verifier to use multi-step trajectory information, while keeping its output interpretable as a value estimate for the candidate action. During candidate generation, the current action slot is first masked to produce a candidate action. To verify a sampled candidate $a_{t,i}$, we insert it into the current action slot and evaluate the value head.
% Sharing the transformer representation allows the verifier to use
% multi-step trajectory information, while keeping its output interpretable as a
% value estimate for the candidate action. 
% During candidate generation the current action slot is masked; to verify a sampled candidate $a_{t,i}$, we re-insert it into the action slot and read the value head, i.e., $Q_\phi(h_t,a_{t,i})$ is an explicit function of the
% context-action pair $(h_t,a_{t,i})$.

% ---------------------------------------------------------------------
\subsection{Hybrid training}\label{sec:method-train}
% ---------------------------------------------------------------------
 
The generator and verifier are trained jointly, but the two objectives serve
different purposes. The masked reconstruction objective teaches the transformer to
model trajectory structure and produce plausible target-conditioned candidates.
The in-sample value objective calibrates the verifier on dataset-supported
context--action pairs, so that generated candidates can be compared by their
predicted long-term values.
\paragraph{Reconstruction objective.}
Let $P_\theta$ denote the conditional distributions induced by the transformer and
its reconstruction heads. The reconstruction loss is
\begin{equation}
\mathcal L_{\mathrm{recon}}(\theta)=
-\sum_{t=0}^{T}\Bigl(\log P_\theta\bigl(R_t^{\mathcal D}\mid M(\tau_t^{K})\bigr)
+\log P_\theta\bigl(s_t\mid M(\tau_t^{K})\bigr)
+\log P_\theta\bigl(a_t\mid M(\tau_t^{K})\bigr)\Bigr),
\label{eq:recon}
\end{equation}
where, for continuous states or actions, the likelihood terms are implemented
as Gaussian likelihoods or mean-squared errors. The random masks encourage the
model to reconstruct trajectory tokens from bidirectional context, which helps it
capture relationships among returns, states, and actions across different parts
of the trajectory. The autoregressive masks preserve the decision-time structure,
where the current action must be predicted from the available context and the
target return. Together, these masking patterns make the generator better suited
for producing useful candidates under a range of requested returns.
 
\paragraph{In-sample value objective.}
The verifier is trained by temporal-difference learning on dataset actions.
For the in-sample TD residual
\begin{equation}
\Delta_t^\phi=r_t+\gamma Q_{\phi}(h_{t+1},a_{t+1})-Q_{\phi}(h_t,a_t),
\end{equation}
we minimize the asymmetric least-squares (expectile) loss
\begin{equation}
\mathcal L_Q(\phi)=\sum_{t=0}^{T-1}L_2^\nu(\Delta_t^\phi),
\qquad
L_2^\nu(u)=\bigl|\nu-\mathbf 1(u<0)\bigr|\,u^2.
\label{eq:vloss}
\end{equation}
With $\nu=0.5$, Eq.\eqref{eq:vloss} is the ordinary squared TD loss; with
$\nu>0.5$, it weights positive residuals more heavily and learns an upper
expectile of the in-sample value distribution~\citep{newey1987asymmetric,kostrikov2022offline}. Since this objective is computed on dataset supported action pairs, it avoids requiring value estimates for arbitrary unsupported actions. This is well matched to the generator, which is also trained from the offline trajectory distribution and therefore tends to produce candidates near the data supported region. Eq.\eqref{eq:vloss} is an in-sample policy evaluation backup: the bootstrap target uses the next dataset action $a_{t+1}$ rather than a maximization over actions. With $\nu>0.5$, the upper expectile biases the estimate toward high-value in-sample continuations, which provides a practical value estimator for reinforced verification. 

% We note that Eq.\eqref{eq:vloss} is an in-sample policy evaluation backup: the bootstrap target uses the next dataset action $a_{t+1}$ rather than a maximization over actions. With $\nu>0.5$ the upper expectile biases the estimate
% toward the in-sample continuations, which is the sense in which $Q_\phi$ approximates the optimal value $Q^*$ of Eq.\eqref{eq:insample}. The bootstrap target is computed with a stop-gradient (a slowly updated target copy of the value head), so that only the current-step value is differentiated, following standard practice for stabilizing temporal-difference learning with function approximation.

% The theoretical analysis below does
% not require this estimator to be globally optimal; it only assumes that
% $Q_\phi$ is accurate on the generated candidate set. The bootstrap target is
% computed with a stop-gradient using a slowly updated target copy of the value
% head, so that only the current-step value is differentiated, following standard
% practice for stabilizing temporal-difference learning with function
% approximation.

\paragraph{Joint objective.}
The final training objective combines the two losses,
\begin{equation}
\mathcal L(\theta,\phi)=\mathcal L_{\mathrm{recon}}(\theta)
+\lambda_Q\,\mathcal L_Q(\phi),
\label{eq:joint}
\end{equation}
where $\lambda_Q$ balances the two terms; we use $\lambda_Q=1$ unless
otherwise specified.
 
% ---------------------------------------------------------------------
\subsection{Verifier-guided inference}
\label{sec:method-infer}
% ---------------------------------------------------------------------
 
At test time, the user specifies a target return $g_t$. A standard
return-conditioned policy would generate one action directly from this target. In
contrast, \ourmethod\ first samples a local set of prompts around $g_t$, generates
one candidate action from each prompt, and then uses the verifier to select the
candidate whose predicted value best matches the original target. The inference
procedure therefore consists of three steps.
 
\paragraph{Step 1: candidate generation.}
Prompts are drawn from a local interval around $g_t$,
\begin{equation}
g_{t,i}\sim\mathrm{Unif}\bigl([g_t-\delta,\,g_t+\delta]\bigr),
\qquad i=1,\ldots,N.
\label{eq:prompt-sample}
\end{equation}
For each $g_{t,i}$, a target-conditioned input is built by replacing the
current return token with $g_{t,i}$ and masking the current action slot; all
$N$ inputs are processed in one batched forward pass,
\begin{equation}
\underbrace{\Bigl\{\bigl(R_{t-K+1}^{\mathcal D},s_{t-K+1},a_{t-K+1},\ldots,
g_{t,i},s_t,\;\rule[-0.5ex]{0.2cm}{0.5pt}\;\bigr)\Bigr\}_{i=1}^{N}}
_{\text{target-conditioned masked inputs}}
\;\longrightarrow\;
\underbrace{\bigl\{(a_{t,i},\,q_{t,i})\bigr\}_{i=1}^{N}}
_{\text{candidates and verifier values}},
\label{eq:parallel}
\end{equation}
which yields the candidate set $\{a_{t,i}\}_{i=1}^{N}$ with
$a_{t,i}\sim\pi_{\mathrm{gen}}(\cdot\mid h_t,g_{t,i})$.
 
\paragraph{Step 2: verification and selection.}
Each candidate is scored by the evaluative pathway,
$q_{t,i}=Q_\phi(h_t,a_{t,i})$, and the executed action is the candidate whose
verified value is closest to the requested target,
\begin{equation}
i^*\in\arg\min_{i\in[N]}\,|q_{t,i}-g_t|,
\qquad
a_t=a_{t,i^*},
\label{eq:select}
\end{equation}
The generator and verifier therefore play complementary roles. The generator
provides multiple plausible actions under nearby prompts, while the verifier
selects the action whose predicted value best aligns with the original target
$g_t$. If the candidate values cover the target region, this rule selects a
candidate close to the requested value. If all candidates fall below a very high
target, the same rule naturally favors the highest-valued candidate among the
generated set.
 
\paragraph{Step 3: target update.}
After executing $a_t$ and observing the reward $r_t$, the requested
return-to-go is updated by
\begin{equation}
g_{t+1}=\frac{g_t-r_t}{\gamma},\qquad \gamma>0,
\label{eq:rtg-update}
\end{equation}
and the procedure repeats at the next timestep. This update carries the remaining
target through the episode: after each action, the next decision is made using the
updated context $h_{t+1}$ and the updated target $g_{t+1}$. 
% A full rollout is
% therefore obtained by repeatedly applying the same procedure. Algorithm~\ref{alg:doctor} summarizes the full procedure.
A full rollout is therefore obtained by repeatedly applying the same procedure. 
% Propositions~\ref{prop:exact}--\ref{prop:robust} and
% Corollary~\ref{cor:coverage} are single-step, value-level statements. 
At rollout
time the rule is applied at every step while the target is propagated by the
return-to-go update~\eqref{eq:rtg-update}, so smaller per-step value gaps push the
remaining return-to-go toward the requested level at each step. We therefore
expect closer episodic alignment as a trend rather than a pointwise guarantee,
since transition stochasticity and the accumulation of per-step verifier error
over the horizon are not controlled by the single-step bounds.
Algorithm~\ref{alg:doctor} summarizes the full procedure.

\begin{algorithm}[t]
\caption{\ourmethod: target-conditioned generation with reinforced
verification}
\label{alg:doctor}
\begin{algorithmic}[1]
\STATE \textbf{Input:} offline dataset $\mathcal D$, transformer parameters
$\theta$, verifier parameters $\phi$, target return $g_0$, bandwidth
$\delta$, sampling budget $N$.
\STATE \textbf{Training:}
\FOR{each gradient step}
\STATE Sample segments $\tau_t^{K}$ from $\mathcal D$ and construct masked
inputs $M(\tau_t^{K})$ as in \eqref{eq:mask}.
\STATE Update $(\theta,\phi)$ by minimizing the joint objective
\eqref{eq:joint}.
\ENDFOR
\STATE \textbf{Inference:}
\STATE Initialize the environment, observe $s_0$, and Initialize the context
$h_0$.
\FOR{$t=0,1,\ldots,T$}
\STATE Sample prompts
$g_{t,i}\sim\mathrm{Unif}([g_t-\delta,g_t+\delta])$, $i=1,\ldots,N$.
\STATE Generate candidates
$a_{t,i}\sim\pi_{\mathrm{gen}}(\cdot\mid h_t,g_{t,i})$ in one batched pass.
\STATE Verify candidates: $q_{t,i}=Q_\phi(h_t,a_{t,i})$.
\STATE Select $i^*\in\arg\min_{i\in[N]}|q_{t,i}-g_t|$ and execute
$a_t=a_{t,i^*}$.
\STATE Observe $r_t,s_{t+1}$; update $g_{t+1}=(g_t-r_t)/\gamma$ and the
context $h_{t+1}$.
\ENDFOR
\end{algorithmic}
\end{algorithm}

% #####################################################################
% \section{Theoretical analysis}
% \label{sec:theory}
% % #####################################################################

\section{Theoretical analysis}
\label{sec:theory}
% #####################################################################

We now analyze the verifier-guided selection step in \ourmethod. This rule is
closely related to best-of-$N$ inference-time alignment and verifier-based
reranking, where multiple candidates are generated and a learned scoring model is
used for selection \citep{beirami2025theoretical,huang2025best}.
Our setting differs from standard best-of-$N$ selection in that the goal is not to
select the candidate with the largest score, but to select the candidate whose
value is closest to a requested target. We therefore analyze alignment through the
attainable value image induced by $Q^*(h,a)$ and the nearest candidate-value gap.

Given a finite set of generated candidate actions, the rule selects the candidate
whose verifier-estimated value is closest to the requested target. The resulting
alignment error is controlled by two factors: how close the generated candidate
set comes to the target value, and how accurately the verifier scores those
candidates.
% We now analyze the verifier-guided selection step in \ourmethod.
% % The goal is
% % not to prove global optimality or exact episodic return matching in every
% % environment. Instead, we focus on the local decision-level mechanism that drives
% % the method. 
% Given a finite set of generated candidate actions, the rule selects
% the candidate whose verifier-estimated value is closest to the requested target. The
% resulting alignment error is controlled by two factors: how close the generated
% candidate set comes to the target value, and how accurately the verifier scores
% those candidates.
Fix a decision context $h$ and target $g$. Let
$\mathcal C_N(h,g)=\{a_i\}_{i=1}^{N}$ be the generated candidate set, let
$q_i=Q_\phi(h,a_i)$, and let
\begin{equation}
i^*\in\arg\min_{i\in[N]}|q_i-g|,
\qquad a^*=a_{i^*}.
\end{equation}
Define the candidate coverage gap
\begin{equation}
\rho_N(h,g)=\min_{i\in[N]}|Q^*(h,a_i)-g|.
\label{eq:rhoN}
\end{equation}
This quantity is the smallest value-level alignment error that could be achieved
by the generated candidate set if the verifier were exact. It separates candidate
generation from value estimation: even a perfect verifier cannot select an action
closer to the target than the best candidate available in the set.
\begin{assumption}[Verifier accuracy on generated candidates]
\label{asm:candidate-accuracy}
For the candidate set $\mathcal C_N(h,g)$ generated at context $h$ and target $g$,
the learned verifier satisfies
\begin{equation}
\max_{a_i\in\mathcal C_N(h,g)}
|Q_\phi(h,a_i)-Q^*(h,a_i)|\le\epsilon
\end{equation}
for some $\epsilon\ge0$.
\end{assumption}
Assumption~\ref{asm:candidate-accuracy} does not require
the verifier to be accurate over the entire action space. It only requires accuracy on the finite candidate set considered at inference. This matches the design of \ourmethod: the generator tends to propose actions near the data-supported region, while in-sample value learning calibrates the verifier on that region.\newline

\noindent\textbf{Exact verification.}
\label{sec:theory-exact}
We first consider the ideal case in which the verifier is exact on the generated
candidate set. This case isolates the role of candidate coverage without the
additional complication of value-estimation error.
\begin{proposition}[Alignment under exact verification]
\label{prop:exact}
Suppose the verifier is exact on the generated candidates, i.e.,
$Q_\phi(h,a_i)=Q^*(h,a_i)$ for all $i\in[N]$. Then the selected action satisfies
\begin{equation}
\mathcal E(h,g;a^*)=\rho_N(h,g).
\end{equation}
Moreover:
\begin{enumerate}
\item[(i)] If $g>\max_{i\in[N]}Q^*(h,a_i)$, then
\begin{equation}
a^*\in\arg\max_{i\in[N]}Q^*(h,a_i).
\end{equation}
Thus, when the requested target is above all values covered by the candidates,
the rule reduces to best-candidate value maximization.
\item[(ii)] If the candidate set contains an action $a_j$ such that
$|Q^*(h,a_j)-g|\le\alpha$ for some $\alpha\ge0$, then
\begin{equation}
\mathcal E(h,g;a^*)\le\alpha.
\end{equation}
In particular, if the candidate values become dense around a feasible target
$g$, the exact-verifier alignment error vanishes with the candidate coverage
gap $\rho_N(h,g)$.
\end{enumerate}
\end{proposition}
Proposition~\ref{prop:exact} provides the basic interpretation of the
verifier-guided selection. If the verifier is exact, selecting the candidate whose
verified value is closest to $g$ is the same as selecting the candidate whose true
value is closest to $g$. Therefore, the remaining error is exactly the coverage
gap of the generated candidate set. The two cases also explain how the same rule
handles different target regimes. When the target is higher than all candidate
values, the nearest-value rule chooses the highest-value candidate. When the
target lies near values covered by the candidates, the rule behaves as a
target-matching mechanism.\newline

\noindent\textbf{Approximate verification.}
\label{sec:theory-robust}
We next consider the practical case in which the verifier is learned and may have
estimation error. The following result shows how this error affects the alignment
guarantee.
\begin{proposition}[Robustness under approximate verification]
\label{prop:robust}
Under Assumption~\ref{asm:candidate-accuracy}, the action selected by
\eqref{eq:verifier-rule} satisfies, for every target $g$,
\begin{equation}
\mathcal E(h,g;a^*)\le \rho_N(h,g)+2\epsilon.
\label{eq:robust-bound}
\end{equation}
\end{proposition}
Proposition~\ref{prop:robust} shows that the learned verifier introduces only an
additive penalty relative to the best candidate available in the generated set.
The first term, $\rho_N(h,g)$, is controlled by candidate generation: the candidate
set must contain an action whose true value is close to the target. The second
term, $2\epsilon$, is controlled by verifier accuracy: even if such a candidate
exists, the verifier must score the candidates accurately enough to select it.
This explains why increasing the candidate budget can help only up to the point
where verifier error becomes the dominant limitation.\newline

\noindent\textbf{Candidate coverage.}
\label{sec:theory-coverage}
The previous two propositions are deterministic statements conditioned on the
generated candidate set. We now describe a standard local coverage condition under
which the candidate coverage gap decreases as the number of generated candidates
increases.
% \begin{corollary}[Coverage consistency]
% \label{cor:coverage}
% Fix a context $h$ and let $g$ be an interior locally covered target. Let
% $Q_{\max}=\sup_{\tilde h,a}|Q^*(\tilde h,a)|<\infty$ and $D=Q_{\max}+|g|$ with
% $D>0$. Suppose the candidate
% values
\begin{corollary}[Coverage consistency] \label{cor:coverage} Fix a context $h$ and a target $g$. Let $Q_{\max}=\sup_{\tilde h,a}|Q^*(\tilde h,a)|<\infty$ and $D=Q_{\max}+|g|$ with $D>0$. Suppose the candidate values
\begin{equation}
X_i=Q^*(h,a_i),\qquad i=1,\ldots,N,
\end{equation}
are i.i.d. conditional on $(h,g,\delta)$ and have a density $f$ satisfying
$f(x)\ge\underline f>0$ on $[g-w_0,g+w_0]$ for some width $w_0\in(0,D]$. Then
\begin{equation}
\mathbb E[\rho_N(h,g)]
\le
\frac{1}{2\underline f N}+D\exp(-2\underline f Nw_0)
=O(1/N).
\label{eq:coverage-bound}
\end{equation}
% Consequently, under Assumption~\ref{asm:candidate-accuracy},
% \begin{equation}
% \limsup_{N\to\infty}\mathbb E[\mathcal E(h,g;a^*)]\le 2\epsilon.
% \end{equation}
% \end{corollary}
Consequently, if Assumption~\ref{asm:candidate-accuracy} holds uniformly with an error bound $\epsilon$ independent of $N$, then \begin{equation} \limsup_{N\to\infty}\mathbb E[\mathcal E(h,g;a^*)]\le 2\epsilon. \end{equation} \end{corollary}
Corollary~\ref{cor:coverage} explains the role of sampling multiple candidates.
If the generator can produce candidate values with non-negligible density around
the requested target, then drawing more candidates increases the chance that at
least one candidate lies close to the target value. As a result, the expected
coverage gap decreases with the sampling budget $N$. 
% However, the second statement
% also shows that candidate sampling alone cannot remove verifier error. As $N$
% grows, the coverage gap can shrink, but the limiting alignment error is still
% controlled by the accuracy of $Q_\phi$ on the generated candidates.
The second statement also shows that candidate sampling alone cannot remove verifier error. As $N$ grows, the coverage gap can shrink, but the limiting alignment error is still controlled by the accuracy of $Q_\phi$ on the generated candidates, provided this accuracy bound remains stable as the candidate set increases.
This result also gives a value-level interpretation of controllability. When the
generator locally covers the requested target region and the verifier is accurate
on the generated candidates, changing the target return changes the value level
toward which the selection rule moves. Thus, controllability arises from the
combination of local candidate coverage and verifier-based selection.

\section{Experiments}
\label{sec:experiments}

We evaluate \ourmethod\ along dimensions that correspond to the main claims
of the paper. Section~\ref{sec:alignment} studies target alignment on
locomotion tasks with reduced high-return coverage, testing whether
the realized episodic return tracks the requested target more closely than
standard return-conditioned sequence models. Section~\ref{sec:double-check-experiment}
examines the mechanisms behind this improvement through component ablations. Section~\ref{sec:return-maximization} evaluates standard D4RL return-maximization
performance to check whether target-aligned control remains competitive with
offline RL and sequence-modeling baselines. Finally, Section~\ref{sec:epicare-control}
uses the simulated EpiCare benchmark to test whether changing the target return can
modulate the operating point of a single trained policy.

\subsection{Setup}
\label{sec:setup}

\begin{itemize}
    \item \textbf{Benchmarks}. We evaluate on D4RL~\citep{fu2020d4rl} and
    EpiCare~\citep{hargrave2024epicare}. For D4RL, we use Gym Locomotion-v2,
    Maze2D-v1, and Adroit-v1 tasks. These tasks test complementary aspects of
    offline control: locomotion performance, stitching of sub-trajectories, and
    high-dimensional dexterous manipulation. EpiCare is a simulated benchmark for
    dynamic treatment regimes. It includes heterogeneous treatment effects, partial
    observability, short treatment horizons, and adverse events, which makes it
    suitable for studying whether target-conditioned policies can change treatment
    aggressiveness. More environment details are provided in Appendix~\ref{app:appenC}.

    \item \textbf{Baselines}. We compare against representative offline RL and
    return-conditioned sequence-modeling baselines. The value-based and actor-critic
    baselines include CQL~\citep{kumar2020conservative}, IQL~\citep{kostrikov2022offline},
    and TD3+BC~\citep{fujimoto2021minimalist}. The supervised or return-conditioned
    baselines include BC, DT~\citep{chen2021decision}, and MTM~\citep{wu2023masked}.
    This comparison separates two effects: whether return-conditioned sequence models
    can follow a requested target, and whether the reinforced verification improves over
    both pure sequence modeling and conventional return-maximizing offline RL.

    \item \textbf{Evaluation protocol}. For alignment evaluation, the
    x-axis is the requested target return and the y-axis is the realized episodic
    return. The ideal alignment curve is the diagonal. Results are reported over five random seeds. For ablations, we report the
    mean absolute error between target and realized returns across multiple target
    levels. In EpiCare, we evaluate policies using 1,000 rollouts and report
    both clinical return and adverse events, since higher return alone does not fully
    characterize clinical utility.
\end{itemize}

\subsection{Empirical target-return alignment}
\label{sec:alignment}

We first evaluate whether \ourmethod\ can make realized rollout returns track
requested target returns. This is the central empirical requirement for
target-conditioned control: a policy should not only generate plausible actions
under a target return, but should also produce outcomes that move toward the
requested return level. We test this property under deliberately difficult data
coverage conditions, where high-return trajectories are partially removed from the
offline dataset.

As a qualitative illustration, Figure~\ref{fig:toy} shows trajectories in
maze2d-umaze-dense under target returns 10, 40, and 80. The reward at each step is
\begin{equation}
    r_t=\exp\left(-\left\|g_{\mathrm{achieved},t}-g_{\mathrm{desired}}\right\|_2\right),
\end{equation}
where $g_{\mathrm{desired}}$ is the goal position and
$g_{\mathrm{achieved},t}$ is the achieved position at time $t$. The episode horizon
is 300. The trajectories show that changing the target return changes the policy
behavior in the intended direction: higher target returns guide the agent closer
to the goal. This toy example is meant as a visual demonstration of controllable
behavior; the following experiments provide the quantitative alignment evaluation.

\begin{figure}
	\centering
    \begin{minipage}{\textwidth}
        \centering
        \includegraphics[width=1.0\textwidth, keepaspectratio]{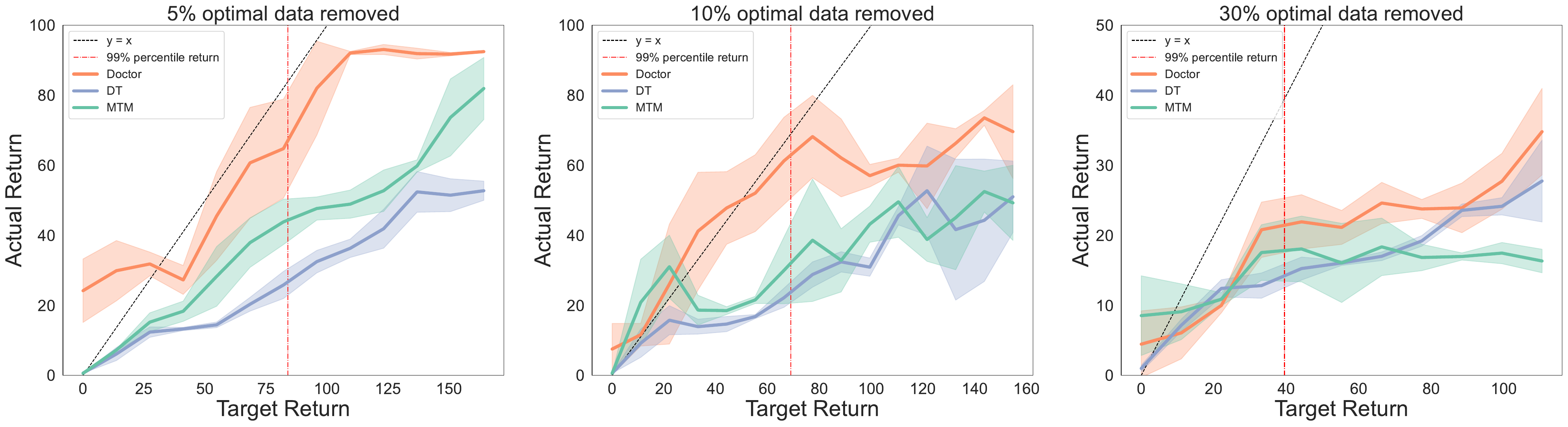}
    \end{minipage}
    \begin{minipage}{\textwidth}
        \centering
        \includegraphics[width=1.0\textwidth, keepaspectratio]{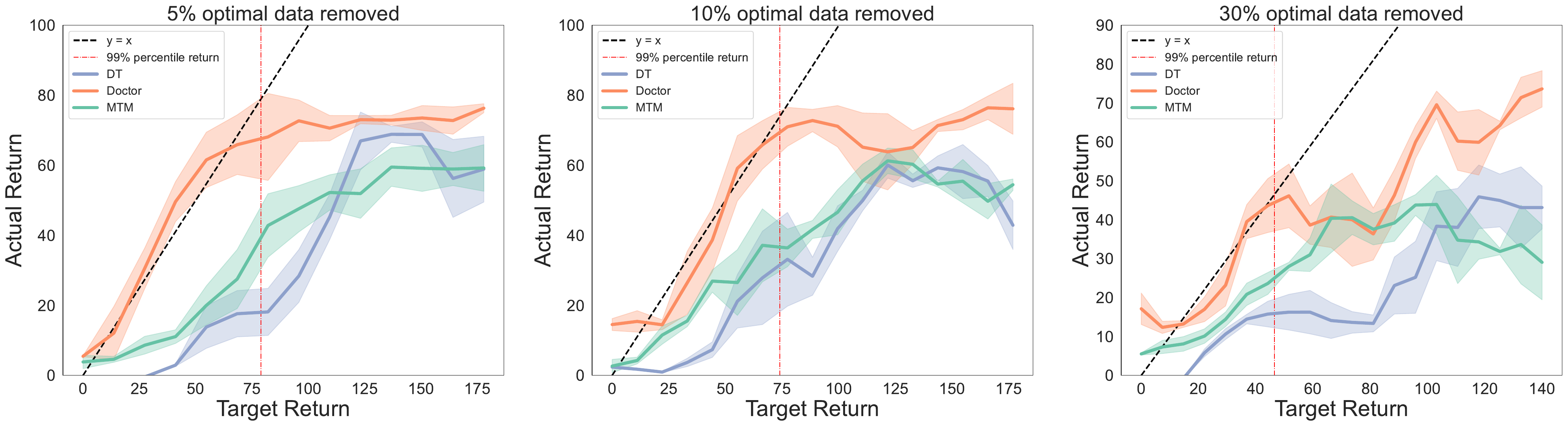}
    \end{minipage}
    \caption{Target-return alignment on hopper-medium-replay-v2 (top) and
    walker2d-medium-replay-v2 (bottom) after removing top-return trajectories from
    the dataset. The x-axis is the requested target return and the y-axis is the
    realized episodic return. The dashed black line is ideal alignment, and the
    dashed red line marks the maximum return remaining in the offline dataset.
    \ourmethod\ stays closer to the ideal diagonal than DT and MTM across a wide
    range of targets, including underrepresented regions and targets near or above
    the dataset maximum.}
    % \vspace{-0.1in}
    \label{fig:alignment}
\end{figure}

We next evaluate hopper-medium-replay-v2 and walker2d-medium-replay-v2 after
removing the top-return trajectories from the dataset. This intervention reduces
coverage of high-quality trajectories and creates underrepresented return regions,
which is precisely where return-conditioned behavioral cloning can fail: the
prompt requests a return level, but the model may not have learned a reliable
mapping from that prompt to actions that realize the requested return.

\begin{wrapfigure}{r}{0.5\textwidth}
    \centering
    % \vspace{-0.15in}
    \includegraphics[width=0.5\textwidth, keepaspectratio]{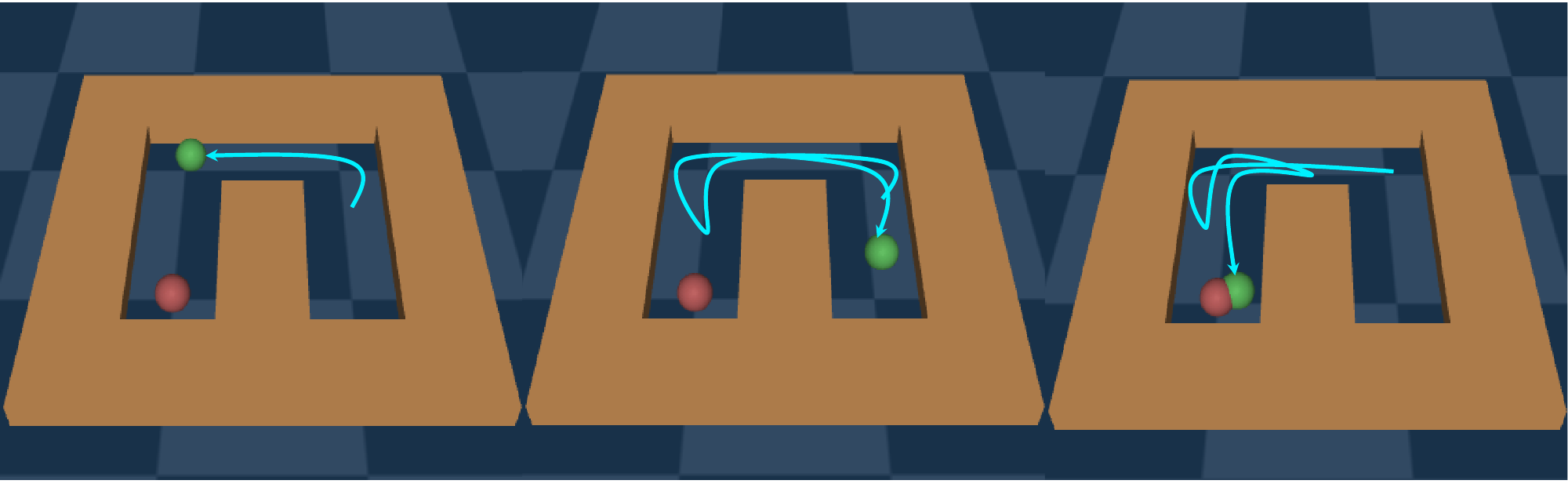}
    \caption{Trajectories generated under three target returns, 10, 40, and 80,
    resulting in realized returns of 37.80, 46.85, and 80.16, respectively. The
    example illustrates that increasing the target return changes the trajectory
    toward higher-return behavior; quantitative target alignment is evaluated in
    Figure~\ref{fig:alignment}.}
    \label{fig:toy}
    % \vspace{-0.15in}
\end{wrapfigure}

Figure~\ref{fig:alignment} reports the results. DT often deviates from the ideal
diagonal when the target lies in underrepresented regions, showing that conditioning
on the target return alone is not sufficient for accurate alignment. MTM benefits
from bidirectional masked sequence modeling and is usually closer to the diagonal
than DT, suggesting that stronger sequence reconstruction improves interpolation.
However, \ourmethod\ is consistently closest to the ideal alignment line. This
provides empirical support for the central claim that verifier-guided
selection helps translate target-conditioned generation into better realized
target alignment. When the requested target exceeds the maximum return observed in
the remaining dataset, \ourmethod\ tends to select high-value candidates among those generated, which is the desired behavior when the target lies above the candidate-covered value range.

\subsection{Mechanism ablations}
\label{sec:double-check-experiment}

We next examine whether the two main mechanisms of \ourmethod\ contribute to
alignment. The sampling budget $N$ controls how many target-conditioned candidates
are available for selection, and the verifier-guided selection step determines whether the
model uses the verifier to choose among those candidates. These ablations therefore
test whether alignment improves when candidate coverage is increased and whether
removing verifier-guided selection harms alignment.

\begin{wrapfigure}{r}{0.5\textwidth}
    \centering
    \vspace{-0.2in}
    \includegraphics[width=0.49\textwidth, keepaspectratio]{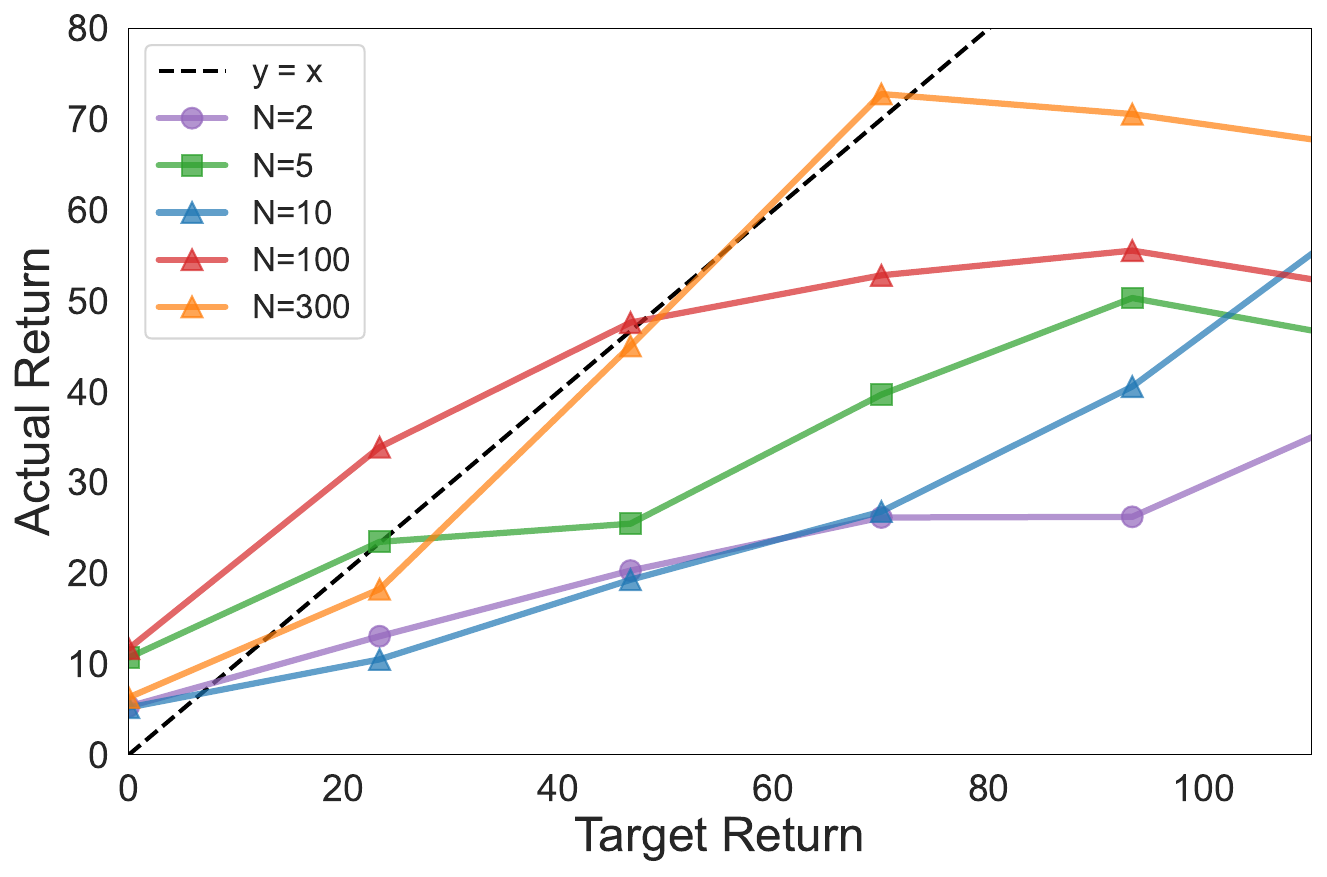}
    \vspace{-0.1in}
    \caption{Effect of the sampling budget $N$ on target-return alignment. Larger $N$
    provides more candidate actions for verifier-based selection and improves
    alignment qualitatively, consistent with the candidate-coverage mechanism in
    Corollary~\ref{cor:coverage}.}
    % \vspace{-0.2in}
    \label{fig:hyperparameter_N}
\end{wrapfigure}

We conduct the candidate-budget ablation on hopper-medium-replay-v2 with 10\% of
the best data removed, using $N\in\{2,5,10,100,300\}$. As shown in
Figure~\ref{fig:hyperparameter_N}, alignment improves as $N$ increases. With a
small budget such as $N=2$, the model behaves similarly to an ordinary
return-conditioned generator: only a few candidates are available, so the verifier
has limited opportunity to correct a misaligned prompt-action mapping. Increasing
$N$ gives the generative pathway more chances to cover the local value region, and
the verifier can then select a candidate whose predicted value better matches the
requested target. This trend is qualitatively consistent with the coverage effect
described in Corollary~\ref{cor:coverage}: a larger candidate set increases the
chance of including a candidate near the target value.

\begin{table}[h]
\centering
\caption{Absolute alignment error in the ablation study. Results are averaged over
three seeds. Lower is better. 'w/o RM' removes the random-masking pattern, and
'w/o RV' removes the reinforced verification.}
\begin{tabular}{@{}lccc@{}}
\toprule
\textbf{Approach} & halfcheetah-M & hopper-M & walker2d-M \\
\midrule
w/o RM & 32.3{\scriptsize$\pm4.6$} & 14.5{\scriptsize$\pm2.4$} & 24.1{\scriptsize$\pm6.6$} \\
w/o RV & 26.2{\scriptsize$\pm5.1$} & 12.9{\scriptsize$\pm2.7$} & 28.6{\scriptsize$\pm7.3$} \\
\textbf{\ourmethod} & \textbf{16.2}{\scriptsize$\pm5.8$} & \textbf{10.4}{\scriptsize$\pm3.3$} & \textbf{12.0}{\scriptsize$\pm8.1$} \\
\bottomrule
\end{tabular}

\label{tab:ablation-error}
\end{table}

\begin{table}[h]
\centering
\caption{Training and inference time comparison. The additional verification cost
of \ourmethod\ is small because candidate generation and value prediction are
computed in a batched forward pass.}
\begin{tabular}{@{}lccc@{}}
\toprule
\textbf{Time Complexity} & DT & MTM & \textbf{\ourmethod} \\ \midrule
Inference (seconds)      & 0.01  & 0.012 & 0.013 \\
Training (seconds)       & 2.13  & 1.29  & 1.35  \\ \bottomrule
\end{tabular}

\label{tab:time}
\vspace{-0.2in}
\end{table}

We also quantify the contribution of the main components in
Table~\ref{tab:ablation-error}. Removing random masking (w/o RM) increases the
alignment error, indicating that masked reconstruction contributes to broad return
coverage and stronger interpolation over trajectory contexts. Removing the
reinforced verification (w/o RV) also degrades alignment, suggesting that merely
producing target-conditioned actions is insufficient. The verifier-based selection
step is important because it chooses among generated candidates according to their
predicted values, rather than executing a single prompt-conditioned action. The
full model achieves the lowest absolute error across all three MuJoCo medium
tasks, providing empirical support for the interpretation that \ourmethod's
alignment benefits from the combination of candidate generation and reinforced
verification.

Finally, Table~\ref{tab:time} shows that the reinforced verification introduces only
minor computational overhead. Although \ourmethod\ samples multiple candidate
prompts, these candidates are processed in parallel, and the value head shares the
same transformer representation as the action head. Thus, the empirical alignment
improvement does not require a large increase in training or inference time.

\subsection{Return maximization performance}
\label{sec:return-maximization}

When a target is set aggressively, the reinforced verification is expected to favor high-value candidates when all generated candidate values fall below the requested target. We therefore evaluate whether \ourmethod\ remains competitive on standard offline RL benchmarks where
the usual objective is to maximize return.

\begin{table*}[t!]
    \centering
    \caption{Offline normalized returns on D4RL Gym locomotion tasks. Higher is
    better. \ourmethod\ achieves the highest aggregate score while retaining the
    ability to condition on different target returns.}
    {\footnotesize
    \begin{tabular}{llllllll}
    \toprule
    \textbf{Locomotion} & BC & CQL & IQL & TD3+BC & DT & MTM & \textbf{\ourmethod} \\
    \midrule
    halfCheetah-MR & 40.2{\scriptsize$\pm 1.2$} & 46.1{\scriptsize$\pm 0.8$} & 45.4{\scriptsize$\pm 1.0$} & 45.6{\scriptsize$\pm 0.9$} & 39.1{\scriptsize$\pm 1.6$} & 43.0{\scriptsize$\pm 1.7$} & 46.6{\scriptsize$\pm 0.9$} \\
    hopper-MR      & 36.6{\scriptsize$\pm 4.6$} & 96.2{\scriptsize$\pm 5.2$} & 93.4{\scriptsize$\pm 4.9$} & 72.6{\scriptsize$\pm 9.4$} & 84.3{\scriptsize$\pm 3.5$} & 93.1{\scriptsize$\pm 3.4$} & 98.8{\scriptsize$\pm 2.1$} \\
    walker2d-MR    & 25.3{\scriptsize$\pm 8.7$} & 73.5{\scriptsize$\pm 1.6$} & 79.5{\scriptsize$\pm 2.3$} & 81.8{\scriptsize$\pm 1.2$} & 66.0{\scriptsize$\pm 3.8$} & 77.3{\scriptsize$\pm 2.9$} & 86.2{\scriptsize$\pm 2.6$} \\
    halfCheetah-M  & 41.8{\scriptsize$\pm 1.1$} & 47.0{\scriptsize$\pm 0.8$} & 48.1{\scriptsize$\pm 0.4$} & 47.2{\scriptsize$\pm 0.6$} & 42.0{\scriptsize$\pm 1.3$} & 44.1{\scriptsize$\pm 0.9$} & 48.4{\scriptsize$\pm 0.7$} \\
    hopper-M       & 55.7{\scriptsize$\pm 5.5$} & 58.5{\scriptsize$\pm 4.8$} & 66.2{\scriptsize$\pm 3.2$} & 60.3{\scriptsize$\pm 6.7$} & 65.2{\scriptsize$\pm 3.8$} & 64.9{\scriptsize$\pm 3.3$} & 85.6{\scriptsize$\pm 7.4$} \\
    walker2d-M     & 63.9{\scriptsize$\pm 5.9$} & 72.9{\scriptsize$\pm 1.5$} & 79.3{\scriptsize$\pm 2.4$} & 84.2{\scriptsize$\pm 1.6$} & 74.5{\scriptsize$\pm 1.3$} & 73.4{\scriptsize$\pm 1.7$} & 81.1{\scriptsize$\pm 2.7$} \\
    halfCheetah-ME & 55.1{\scriptsize$\pm 1.2$} & 91.2{\scriptsize$\pm 1.7$} & 89.6{\scriptsize$\pm 2.9$} & 92.5{\scriptsize$\pm 1.5$} & 86.8{\scriptsize$\pm 1.7$} & 94.7{\scriptsize$\pm 0.9$} & 95.8{\scriptsize$\pm 0.5$} \\
    hopper-ME      & 53.2{\scriptsize$\pm 4.1$} & 99.5{\scriptsize$\pm 10.3$} & 106.7{\scriptsize$\pm 6.6$} & 101.2{\scriptsize$\pm 8.8$} & 110.6{\scriptsize$\pm 1.2$} & 112.4{\scriptsize$\pm 0.8$} & 113.5{\scriptsize$\pm 0.6$} \\
    walker2d-ME    & 100.8{\scriptsize$\pm 7.6$} & 109.9{\scriptsize$\pm 1.5$} & 109.6{\scriptsize$\pm 2.2$} & 111.5{\scriptsize$\pm 1.3$} & 108.3{\scriptsize$\pm 1.9$} & 111.4{\scriptsize$\pm 1.1$} & 113.7{\scriptsize$\pm 1.3$} \\
    \midrule
    \textbf{Sum} & 472.6 & 694.8 & 717.8 & 696.9 & 676.8 & 714.3 & \textbf{769.7} \\
    \bottomrule
    \end{tabular}
    }

    % \vspace{-0.1in}
    \label{tab:main_results}
\end{table*}

As shown in Table~\ref{tab:main_results}, \ourmethod\ achieves the highest
aggregate score on the locomotion suite and obtains the best result in eight of
nine tasks. On medium-replay and medium datasets, the improvement over DT and MTM
suggests that the verifier contributes useful value-based selection beyond pure
supervised sequence modeling. On medium-expert datasets, \ourmethod\ remains
competitive with strong value-based methods while benefiting from the stability of
trajectory reconstruction. These results suggest that target alignment and return
maximization can be compatible in \ourmethod: the verifier used for alignment can
also help identify high-value candidates when aggressive targets are requested.

\begin{table*}[h]
    \centering
    \caption{Offline normalized returns on Adroit-cloned tasks. Higher is better.
    \ourmethod\ achieves the highest aggregate score and remains competitive on
    high-dimensional dexterous manipulation.}
    \begin{tabular}{llllllll}
    \toprule
    \textbf{Adroit} & BC & CQL & IQL & TD3+BC & DT & MTM & \textbf{\ourmethod} \\
    \midrule
    pen    & 56.1{\scriptsize$\pm 13.1$} & 5.4{\scriptsize$\pm 8.7$} & 83.3{\scriptsize$\pm 7.3$} & 5.1{\scriptsize$\pm 4.9$} & 65.2{\scriptsize$\pm 3.6$} & 80.5{\scriptsize$\pm 4.2$} & 85.7{\scriptsize$\pm 8.5$} \\
    hammer & 0.5{\scriptsize$\pm 0.4$}  & 1.9{\scriptsize$\pm 0.5$} & 3.2{\scriptsize$\pm 0.6$} & 0.1{\scriptsize$\pm 0.3$} & 2.0{\scriptsize$\pm 0.8$} & 5.5{\scriptsize$\pm 1.1$} & 4.9{\scriptsize$\pm 3.1$} \\
    door   & 0.0{\scriptsize$\pm 0.0$}  & 0.5{\scriptsize$\pm 0.2$} & 1.8{\scriptsize$\pm 1.4$} & 0.2{\scriptsize$\pm 0.2$} & 7.9{\scriptsize$\pm 2.2$} & 10.5{\scriptsize$\pm 2.0$} & 9.6{\scriptsize$\pm 3.2$} \\
    \midrule
    \textbf{Sum} & 56.6 & 7.8 & 88.3 & 5.4 & 75.1 & 96.5 & \textbf{100.2} \\
    \bottomrule
    \end{tabular}

    \label{tab:adroit results}
\end{table*}

\begin{table*}[h]
    \centering
    \caption{Offline normalized returns on Maze2D tasks. Higher is better.
    \ourmethod\ achieves the highest aggregate score, indicating strong stitching
    ability in navigation tasks with sub-trajectory composition.}
    \begin{tabular}{llllllll}
    \toprule
    \textbf{Maze2D} & BC & CQL & IQL & TD3+BC & DT & MTM & \textbf{\ourmethod} \\
    \midrule
    umaze  & 8.5{\scriptsize$\pm 7.4$} & 96.0{\scriptsize$\pm 3.7$} & 44.2{\scriptsize$\pm 2.9$} & 29.8{\scriptsize$\pm 11.0$} & 19.0{\scriptsize$\pm 21.6$} & 26.4{\scriptsize$\pm 12.1$} & 97.2{\scriptsize$\pm 9.8$} \\
    medium & 7.1{\scriptsize$\pm 4.5$} & 80.4{\scriptsize$\pm 5.6$} & 32.8{\scriptsize$\pm 3.9$} & 62.1{\scriptsize$\pm 12.2$} & 28.5{\scriptsize$\pm 15.7$} & 20.9{\scriptsize$\pm 7.9$} & 106.2{\scriptsize$\pm 7.3$} \\
    large  & 2.8{\scriptsize$\pm 0.6$} & 52.1{\scriptsize$\pm 12.5$} & 63.0{\scriptsize$\pm 2.9$} & 97.6{\scriptsize$\pm 13.1$} & 30.4{\scriptsize$\pm 22.4$} & 41.7{\scriptsize$\pm 8.0$} & 75.7{\scriptsize$\pm 6.4$} \\
    \midrule
    \textbf{Sum} & 18.4 & 228.5 & 140.0 & 189.5 & 77.9 & 89.0 & \textbf{279.1} \\
    \bottomrule
    \end{tabular}

    \label{tab:maze2d results}
\end{table*}

We further evaluate \ourmethod\ on Adroit and Maze2D. In Adroit, the model must
handle high-dimensional dexterous manipulation, where pure return-conditioned
sequence modeling can be brittle. Table~\ref{tab:adroit results} shows that
\ourmethod\ obtains the highest aggregate score and the best result on Pen, while
remaining competitive on Hammer and Door. In Maze2D, success requires stitching
short sub-trajectories into coherent goal-reaching behavior. Table~\ref{tab:maze2d results}
shows that \ourmethod\ achieves the highest aggregate score, outperforming DT and
MTM by a large margin. These results support the contribution that reinforced
verification improves not only target alignment but also downstream control: the
verifier helps the model select actions that compose trajectories effectively in
environments where long-horizon stitching is important.

\subsection{Clinical policy control in EpiCare}
\label{sec:epicare-control}

Finally, we test whether target-conditioned alignment can support controllable
policy modulation in a simulated clinical setting. This experiment evaluates
whether changing the requested target return moves a single trained policy between
clinically meaningful operating points. It should not be interpreted as a
verification that the local coverage and verifier-accuracy assumptions hold for
every feasible target, nor as evidence from real clinical deployment.

EpiCare provides datasets generated by distinct behavior policies across eight
environments, corresponding to environment seeds 1--8 with different simulated
disease characteristics. We use the Standard-of-Care (SoC) dataset, which emulates
clinician behavior that avoids high-risk treatments based on symptom thresholds.
Because EpiCare has discrete treatments, we adapt baseline methods by optimizing
logits of one-hot encoded action outputs. We evaluate models over 1,000 rollouts
and report both clinical returns and adverse events. Clinical return measures the
overall therapeutic benefit of a policy, whereas adverse events measure unsafe
outcomes; a controllable policy should therefore be able to move between more
conservative and more aggressive treatment regimes.

\begin{wrapfigure}{r}{0.5\textwidth}
    \centering
    % \vspace{-0.2in}
    \includegraphics[width=0.53\textwidth, keepaspectratio]{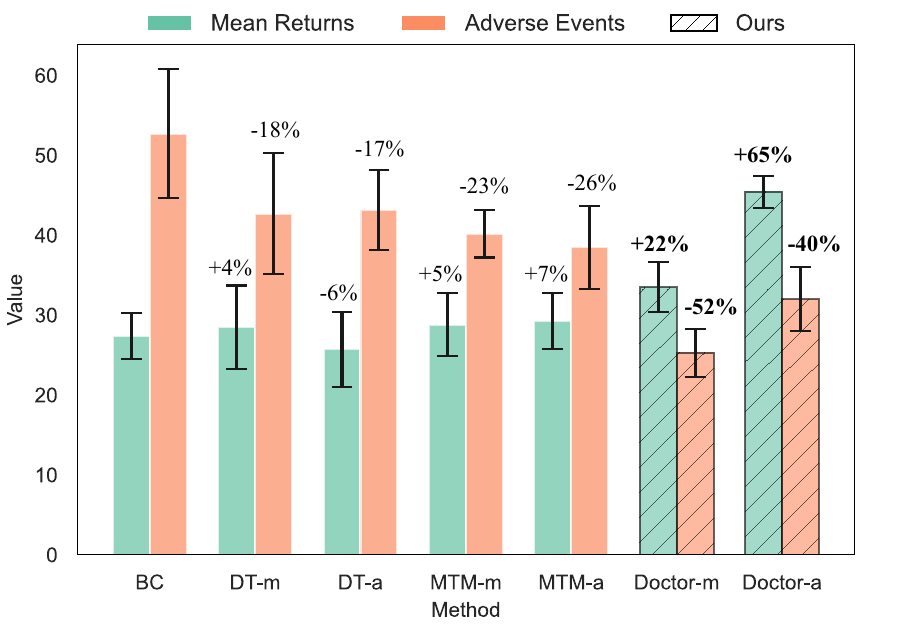}
    
    \caption{Mean returns and adverse-event rates per 10k episodes under moderate
    (m, $0.4\times$) and aggressive (a, $0.8\times$) target returns. Numbers denote
    improvement or decrease compared with BC. \ourmethod\ responds more clearly to
    target changes than DT and MTM, yielding a conservative/aggressive trade-off in
    the simulated EpiCare benchmark.}
    % \vspace{-0.2in}
    \label{fig:care_return}
\end{wrapfigure}

Figure~\ref{fig:care_return} reports clinical return and adverse events per 10k
episodes under moderate ($0.4\times$ maximum dataset return) and aggressive
($0.8\times$ maximum dataset return) targets. DT and MTM show only marginal shifts
relative to BC, indicating that changing the target return alone does not reliably
modulate clinical behavior. By contrast, \ourmethod\ improves returns by
22\%/65\% and reduces adverse events by 52\%/40\% at moderate/aggressive targets
relative to BC. More importantly, changing the target changes the operating point:
the aggressive target increases return while accepting more adverse-event risk,
whereas the moderate target yields a more conservative policy with lower adverse
events. This provides empirical support for the practical claim that the
verifier-guided alignment mechanism can expose a controllable policy knob in a
single trained model.

\begin{figure}
	\centering
	\includegraphics[width=0.6\textwidth]{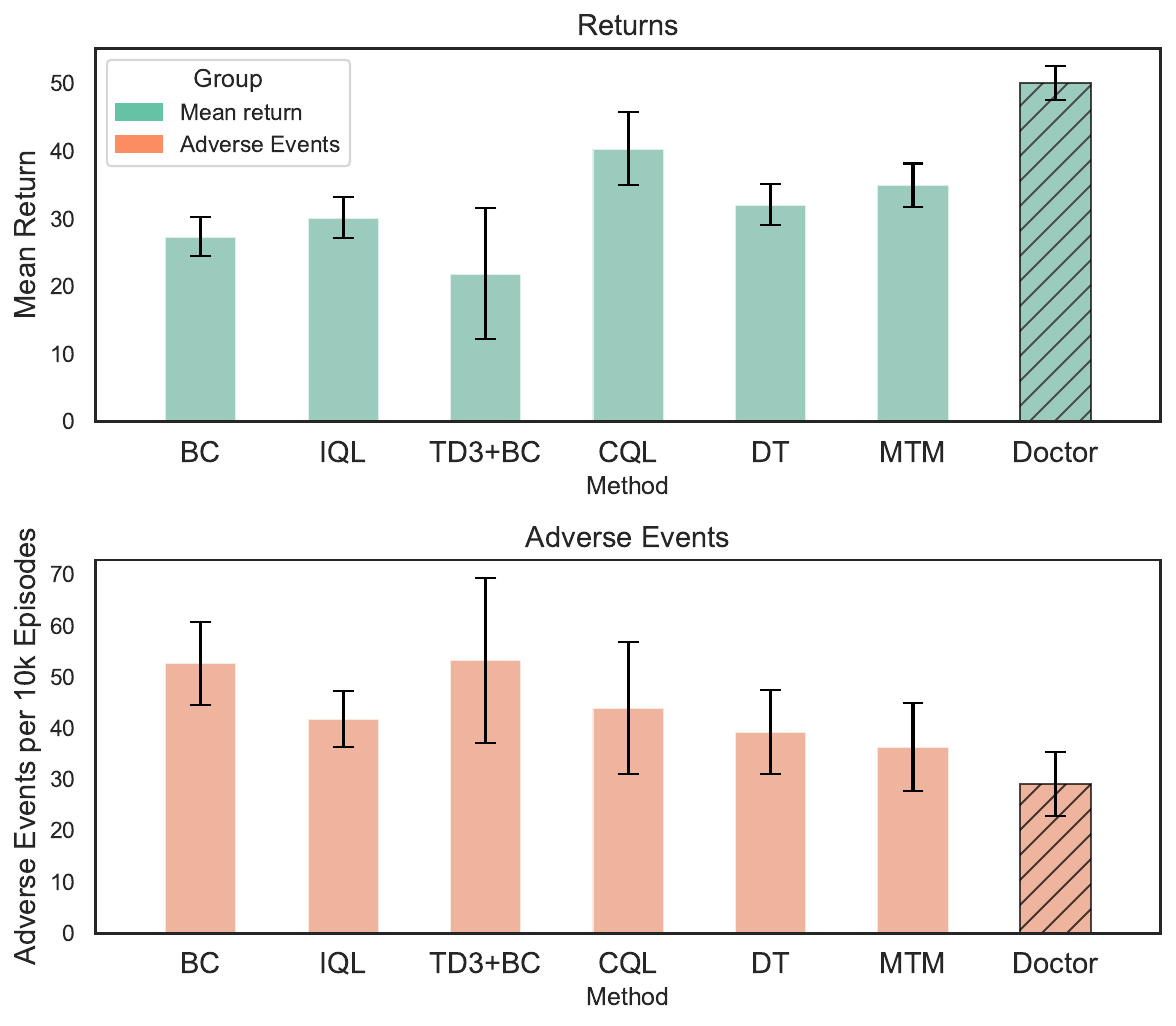}
    \centering
	\caption{Mean returns and adverse events per 10k episodes on EpiCare. Results are averaged over eight structurally distinct simulated environments (environment seeds 1-8) on the SoC dataset. Error bars denote standard deviation. Lower adverse-event rate is better.}
	\label{fig:care_add}
\end{figure}

Figure~\ref{fig:care_add} further compares \ourmethod\ with offline RL baselines on
the SoC dataset across the eight EpiCare environments. \ourmethod\ achieves strong
clinical returns while maintaining the lowest adverse-event rate among the compared
methods. This result suggests that improved alignment is not merely a cosmetic
property of return-conditioned models: in this simulated treatment benchmark, it
corresponds to more effective and safer policy modulation.

\section{Related Work}
\label{sec:related-work}

This section reviews prior work related to \ourmethod, focusing on offline value
learning, sequence modeling for offline decision making, and controllable target
alignment.

\noindent \textbf{Value function learning in offline RL.}
Offline reinforcement learning learns policies from a fixed dataset collected by
one or more behavior policies, without further interaction with the environment.
A central challenge is distributional shift: a learned policy may query actions
that are poorly covered by the dataset, making value estimates unreliable. This
issue has motivated conservative, constrained, and in-sample temporal-difference
learning methods~\citep{sutton2018reinforcement,kumar2019stabilizing,
kumar2020conservative,fujimoto2021minimalist,wu2022supported,
kostrikov2022offline,zhang2023replay,zhang2024exploiting}. CQL~\citep{kumar2020conservative}
regularizes the learned $Q$ function to penalize overestimated
out-of-distribution actions, while TD3+BC~\citep{fujimoto2021minimalist}
combines actor-critic learning with behavior-cloning regularization. IQL~\citep{kostrikov2022offline}
avoids directly maximizing over unseen actions by estimating in-sample values
with expectile regression. Recent studies have also
examined complementary offline-RL issues, including adaptive pessimism through
target Q-values, out-of-distribution Q-value reduction, result-constrained
behavior cloning, diffusion-based policy distillation, and in-sample value
regularization~\citep{liu2024adaptive,wang2025efficient,an2025result,
zhang2025diffusion,liu2026mitigating}. These methods are primarily designed to
extract high-return policies from offline data. In contrast, our goal is
controllable target alignment: the policy should not only perform well, but also
adjust its behavior to match different requested return targets. \ourmethod\ uses
in-sample value learning not as a standalone return-maximization policy, but as a
verifier for candidate actions generated by a target-conditioned sequence model.

\noindent \textbf{Sequence modeling for offline decision making.}
A second line of work formulates offline RL as supervised sequence modeling,
often referred to as reinforcement learning via supervised learning. These
methods avoid explicit bootstrapping during policy learning and benefit from the
stability and scalability of supervised learning. Decision
Transformer~\citep{chen2021decision} trains a Transformer to autoregressively
predict actions conditioned on past trajectory tokens and a desired
return-to-go. Related approaches condition policies on goals, returns, or
return-like prompts and use the prompt as an interface for controlling learned
behavior~\citep{emmons2021rvs,janner2021offline,liu2023emergent,
wu2023elastic,chebotar2023q}. However, return-conditioned supervised learning
depends strongly on data coverage, and changing the target return does not by
itself guarantee that the realized return will match the requested
target~\citep{brandfonbrener2022does,tanaka2024return}. Recent work has therefore
extended return-conditioned sequence models with richer trajectory modeling or
value information. Masked Trajectory Modeling (MTM)~\citep{wu2023masked}
reconstructs missing return, state, and action tokens from partially observed
trajectories, enabling bidirectional representation learning while preserving
decision-time prediction through autoregressive masks. Other methods incorporate
dynamic programming, critic information, or value-guided objectives into
Transformer-based offline RL to improve stitching and recover stronger
return-maximizing policies~\citep{yamagata2023q,gao2024act,wang2024critic,
kim2024adaptive}. These approaches demonstrate that value information can improve
sequence-modeling policies, but their primary objective is usually to improve
final performance or policy optimality. \ourmethod\ instead uses the value head
as a verifier that calibrates controllability: among multiple target-conditioned
candidates, it selects the action whose predicted value is closest to the
requested target.

\noindent \textbf{Controllable target alignment and verifier-guided selection.}
Many real decision-making problems require policies that can operate at different
performance levels, risk preferences, or intervention strengths, rather than a
single uniformly aggressive policy. Such controllability is important in domains
such as clinical decision-making, education, game AI, and robotic
control~\citep{goetz2018personalized,hargrave2024epicare,
singla2021reinforcement,ALAWWAD2025111332,jeon2023raidenv}. Deep reinforcement
learning more broadly provides a neural framework for learning sequential
decision policies from high-dimensional representations and learned predictive
models~\citep{matsuo2022deep}. Return-conditioned models provide a convenient
interface for controllability, since the target return can be changed at
inference time. Nevertheless, prompt-based control alone may be insufficient when
the model has no mechanism to verify whether the generated action is
value-consistent with the requested target. Our inference procedure is related to
verifier-based reranking and best-of-$N$ inference-time alignment, where multiple
candidates are generated and a learned scoring model is used to select among
them~\citep{beirami2025theoretical,huang2025best}. Standard
best-of-$N$ selection typically chooses the highest-scoring candidate, whereas
\ourmethod\ selects the candidate whose verifier value is closest to the
requested target. This changes the role of the verifier from pure reward
maximization to target matching. 
% By combining masked target-conditioned
% generation with verifier-guided selection, \ourmethod\ directly addresses the
% target-alignment problem while retaining competitive offline control performance.

\section{Discussion}

The results suggest that controllability in offline reinforcement learning benefits from separating target-conditioned generation from value-based verification. Rather than relying on a target return as a direct control signal, \ourmethod\ treats it as a proposal mechanism and uses a learned verifier to select actions whose predicted consequences better match the requested target. This design may be useful in safety-sensitive domains where a single high-performing policy is not sufficient and users may need to adjust behavior according to risk tolerance, resource constraints, or intervention intensity. In robotics, such control could support policies that adapt smoothly across performance levels; in healthcare, it could help explore treatment strategies that balance expected benefit and adverse-event risk. At the same time, the current medical evaluation is conducted in simulation, and deployment on real clinical data would require careful validation of verifier calibration, robustness under distribution shift, and domain-specific safety constraints. Future work may also extend the framework beyond scalar return targets toward richer forms of controllable decision making, such as multi-objective, constrained, or preference-conditioned targets.

\section{Conclusion}

This paper introduced \ourmethod, a hybrid sequence modeling and reinforced verification framework for controllable offline decision making. By combining masked trajectory reconstruction with in-sample value learning, \ourmethod\ generates multiple target-conditioned candidate actions and selects the one whose verified value is closest to the requested target. The theoretical analysis shows that the alignment error of this selection rule depends on candidate-value coverage and verifier accuracy, explaining why larger candidate budgets can improve target alignment until value-estimation error becomes limiting. Experiments on D4RL and EpiCare demonstrate that \ourmethod\ improves target alignment in challenging offline settings with reduced high-return coverage, while remaining competitive on standard return-maximization benchmarks. The EpiCare results further show that varying the target return can modulate a single policy between conservative and aggressive operating points, highlighting the potential of reinforced verification for controllable policy behavior in safety-sensitive sequential decision problems.

% \clearpage
\appendix
\section*{Appendix}
This appendix provides supplementary material for \ourmethod. Appendix~\ref{app:proofs}
contains the proofs for the decision-level analysis in Section~\ref{sec:theory}.
Appendix~\ref{app:design-discussion} discusses the architectural design choices
behind target-conditioned generation and reinforced verification. Appendix~\ref{app:env-details}
describes the D4RL and EpiCare environments used in the experiments.
% % Appendix~\ref{app:additional-results}
% and provides additional qualitative and exploratory results. 
Appendix~\ref{app:model-training-details} reports implementation and hyperparameter details.

% #####################################################################
\section{Proofs}
\label{app:proofs}
% #####################################################################

We use the notation of Section~\ref{sec:theory}. In particular,
$q_i=Q_\phi(h,a_i)$, $a^*=a_{i^*}$ with
$i^*\in\arg\min_{i\in[N]}|q_i-g|$, and
$\rho_N(h,g)=\min_i|Q^*(h,a_i)-g|$.

\subsection*{Proof of Proposition~\ref{prop:exact}}
Assume $Q_\phi(h,a_i)=Q^*(h,a_i)$ for every generated candidate. Then the selection
rule becomes
\begin{equation}
 i^*\in\arg\min_{i\in[N]}|Q^*(h,a_i)-g|.
\end{equation}
Therefore,
\begin{equation}
\mathcal E(h,g;a^*)
=|Q^*(h,a^*)-g|
=\min_{i\in[N]}|Q^*(h,a_i)-g|
=\rho_N(h,g),
\end{equation}
which proves the equality.

For item (i), if $g>\max_i Q^*(h,a_i)$, then
$Q^*(h,a_i)-g<0$ for every $i$, and hence
\begin{equation}
 |Q^*(h,a_i)-g|=g-Q^*(h,a_i).
\end{equation}
The function $x\mapsto g-x$ is strictly decreasing, so minimizing
$g-Q^*(h,a_i)$ over the candidate set is equivalent to maximizing
$Q^*(h,a_i)$. Thus,
\begin{equation}
 a^*\in\arg\max_{i\in[N]}Q^*(h,a_i).
\end{equation}

For item (ii), suppose there exists a candidate $a_j$ such that
$|Q^*(h,a_j)-g|\le\alpha$ for some $\alpha\ge0$. By the equality already proved,
\begin{equation}
\mathcal E(h,g;a^*)
=\min_{i\in[N]}|Q^*(h,a_i)-g|
\le |Q^*(h,a_j)-g|
\le\alpha.
\end{equation}
This completes the proof. \hfill$\square$

\subsection*{Proof of Proposition~\ref{prop:robust}}
Let $a^\circ$ be a candidate attaining the coverage gap,
\begin{equation}
 |Q^*(h,a^\circ)-g|=\rho_N(h,g).
\end{equation}
Let $q_* = Q_\phi(h,a^*)$ and $q_\circ=Q_\phi(h,a^\circ)$. By the selection rule,
\begin{equation}
 |q_*-g|\le |q_\circ-g|.
\end{equation}
Using the triangle inequality and Assumption~\ref{asm:candidate-accuracy},
\begin{align}
 |Q^*(h,a^*)-g|
&\le |Q^*(h,a^*)-q_*|+|q_*-g| \\
&\le \epsilon+|q_*-g| \\
&\le \epsilon+|q_\circ-g| \\
&\le \epsilon+|q_\circ-Q^*(h,a^\circ)|+|Q^*(h,a^\circ)-g| \\
&\le \epsilon+\epsilon+\rho_N(h,g) \\
&=\rho_N(h,g)+2\epsilon .
\end{align}
This proves the claim. \hfill$\square$

\subsection*{Proof of Corollary~\ref{cor:coverage}}
Write $X_i=Q^*(h,a_i)$. By assumption, $X_i$ are i.i.d. and have density $f$ satisfying
$f(x)\ge\underline f$ on $[g-w_0,g+w_0]$ for a width $w_0\in(0,D]$. Let
\begin{equation}
 \rho_N(h,g)=\min_{i\in[N]}|X_i-g|.
\end{equation}
Since $|X_i|\le Q_{\max}$, we have
\begin{equation}
0\le \rho_N(h,g)\le D:=Q_{\max}+|g|
\end{equation}
almost surely. Therefore $\Pr(\rho_N(h,g)>u)=0$ for all $u>D$.

For $u\ge0$, define
\begin{equation}
 p_u=\Pr(|X_1-g|<u).
\end{equation}
By independence,
\begin{equation}
 \Pr(\rho_N(h,g)\ge u)=(1-p_u)^N\le \exp(-Np_u).
\end{equation}
For $0\le u\le w_0$, the interval $[g-u,g+u]$ lies inside the local coverage
region, so
\begin{equation}
 p_u=\Pr(|X_1-g|<u)
=\int_{g-u}^{g+u}f(x)\,dx
\ge 2\underline f u.
\end{equation}
Thus,
\begin{equation}
 \Pr(\rho_N(h,g)\ge u)\le \exp(-2\underline f N u),
 \qquad 0\le u\le w_0.
\end{equation}
Using the tail-integral identity for nonnegative random variables and
$w_0\le D$,
\begin{align}
\mathbb E[\rho_N(h,g)]
&=\int_0^\infty \Pr(\rho_N(h,g)\ge u)\,du \\
&=\int_0^{w_0}\Pr(\rho_N(h,g)\ge u)\,du
+\int_{w_0}^{D}\Pr(\rho_N(h,g)\ge u)\,du .
\end{align}
For the first term,
\begin{equation}
\int_0^{w_0}\Pr(\rho_N(h,g)\ge u)\,du
\le \int_0^{w_0}\exp(-2\underline f N u)\,du
\le \frac{1}{2\underline f N}.
\end{equation}
For the second term, monotonicity of the tail probability gives
\begin{equation}
\int_{w_0}^{D}\Pr(\rho_N(h,g)\ge u)\,du
\le D\Pr(\rho_N(h,g)\ge w_0)
\le D\exp(-2\underline f Nw_0).
\end{equation}
Combining the two bounds yields
\begin{equation}
\mathbb E[\rho_N(h,g)]
\le \frac{1}{2\underline f N}+D\exp(-2\underline f Nw_0).
\end{equation}
% The first term is $O(1/N)$ and the second term decays exponentially. Taking
% expectations in Proposition~\ref{prop:robust} gives
% \begin{equation}
% \mathbb E[\mathcal E(h,g;a^*)]
% \le \mathbb E[\rho_N(h,g)]+2\epsilon,
% \end{equation}
% and hence
% \begin{equation}
% \limsup_{N\to\infty}\mathbb E[\mathcal E(h,g;a^*)]\le2\epsilon.
% \end{equation}
% This completes the proof. \hfill$\square$
The first term is $O(1/N)$ and the second term decays exponentially. Therefore,
$\mathbb E[\rho_N(h,g)]\to 0$ as $N\to\infty$. If
Assumption~\ref{asm:candidate-accuracy} holds uniformly with an error bound
$\epsilon$ independent of $N$, then taking expectations in
Proposition~\ref{prop:robust} gives
\begin{equation}
\mathbb E[\mathcal E(h,g;a^*)]
\le \mathbb E[\rho_N(h,g)]+2\epsilon.
\end{equation}
Taking the limit superior yields
\begin{equation}
\limsup_{N\to\infty}\mathbb E[\mathcal E(h,g;a^*)]\le2\epsilon.
\end{equation}
This completes the proof. \hfill$\square$

% #####################################################################
\section{Additional discussion of design choices}
\label{app:design-discussion}
% #####################################################################

\paragraph{Bidirectional masked trajectory modeling.}
Early return-conditioned sequence models such as Decision Transformer and
Trajectory Transformer rely on causal sequence modeling, where each token is
predicted from previous tokens~\citep{chen2021decision,janner2021offline}. In
contrast, masked trajectory modeling uses bidirectional context and trains the
model to reconstruct missing return, state, and action tokens from partially
observed trajectories~\citep{wu2023masked}. Random masks encourage the model to use
information from both past and future tokens during training, while autoregressive
masks preserve the ability to predict actions at inference. This design is useful
for target-conditioned control because it improves interpolation over trajectory
segments associated with different return levels. In the main experiments,
MTM already improves over DT in several underrepresented-return regimes, and
\ourmethod\ builds on this representation by adding reinforced verification.

\paragraph{In-sample value verification.}
The value head in \ourmethod\ is trained with an expectile-style temporal-difference
loss inspired by in-sample value learning~\citep{newey1987asymmetric,kostrikov2022offline}.
For $\nu>0.5$, the asymmetric loss down-weights negative TD residuals and pushes
the learned value toward an upper expectile of the in-sample value distribution. The analysis does not require $Q_\phi$ to exactly solve the maximization in
Eq.~\eqref{eq:insample}; it only requires candidate-level accuracy of $Q_\phi$
on the generated action set, as stated in Assumption~\ref{asm:candidate-accuracy}.
We use this objective as a practical way to calibrate the verifier on
trajectory-supported context--action pairs. The role of the value head is not to
replace the generative policy with a conventional maximizing Q-policy. Instead, it
serves as a verifier that checks whether a generated candidate action is
value-consistent with the requested target. This interpretation matches the
analysis in Section~\ref{sec:theory}: the reinforced verification is useful when the
candidate set covers the target value region and the verifier is accurate on those
candidates. The theory therefore treats verifier error as an explicit residual
term rather than assuming globally perfect value estimation.

% #####################################################################
\section{Environment details}
\label{app:env-details}
\label{app:appenC}
% #####################################################################

\subsection{D4RL}

\paragraph{Gym Locomotion.}
The D4RL Gym locomotion benchmark~\citep{fu2020d4rl} includes continuous-control
tasks based on HalfCheetah, Hopper, and Walker2d. We evaluate Medium-Replay,
Medium, and Medium-Expert datasets. The Medium datasets are collected from
partially trained policies, Medium-Replay datasets contain replay buffers from
training agents, and Medium-Expert datasets mix medium-quality and expert
trajectories. These tasks test whether an offline method can learn robust motor
control from datasets with different quality and coverage. For example, Walker2d
requires the agent to coordinate two legs to move forward while maintaining
balance, making it sensitive to both control quality and data coverage.

\paragraph{Adroit.}
Adroit contains high-dimensional dexterous manipulation tasks involving a
five-fingered robotic hand. We use the Pen, Door, and Hammer tasks. The cloned
datasets mix expert demonstrations with trajectories generated by a behavior
cloning policy. These tasks are challenging for pure return-conditioned sequence
models because they require precise manipulation under high-dimensional action
spaces and imperfect behavior data.

\paragraph{Maze2D.}
Maze2D is a navigation benchmark in which the agent must reach a fixed target
position. We use umaze, medium, and large layouts. These tasks are designed to test
trajectory stitching: useful sub-trajectories may exist in the dataset, but the
learner must compose them into a coherent path to the goal. This makes Maze2D a
natural setting for evaluating whether the verifier helps select actions that
support long-horizon composition rather than only imitating local behavior.

\subsection{EpiCare}

\paragraph{Environment.}
EpiCare~\citep{hargrave2024epicare} is a simulated benchmark for offline
reinforcement learning in dynamic treatment regimes. It formulates longitudinal
patient care as a finite-horizon partially observable Markov decision process. The
hidden state space
$S=\{s_r,s_a,s_1,\ldots,s_{n_s}\}$ contains $n_s$ disease states, a terminal
remission state $s_r$, and a terminal adverse-event state $s_a$. The discrete
action set $A=\{a_1,\ldots,a_{n_a}\}$ represents treatments, and the observation
space $O=[0,1]^{d_o}$ records symptom measurements visible to the clinician.

Disease progression is governed by a transition matrix that is multiplicatively
modulated by the treatment vector $m_a$, together with treatment-specific
remission probabilities $T(s_r\mid s_i,a)$ and adverse-event probabilities
$T(s_a\mid s_i,a)$. For non-terminal disease states, the transition probability is
\begin{equation}
T(s_j \mid s_i,a)=
\left(1-T(s_r \mid s_i,a)-T(s_a \mid s_i,a)\right)
\frac{(m_a)_j T_{i,j}}{\sum_k (m_a)_k T_{i,k}} .
\end{equation}
Observations are generated by applying a treatment-specific symptom shift
$\delta_a$ to a state-dependent Gaussian draw $\tilde{o}$, followed by a logistic
transformation,
\begin{equation}
 o=\operatorname{expit}(\tilde{o}+\delta_a)\in[0,1]^{d_o}.
\end{equation}
The reward function is
\begin{equation}
R(s,a,o)=
\begin{cases}
 r_r, & s=s_r,\\
 r_a, & s=s_a,\\
 -c_a-c_o\sum_i o_i, & \text{otherwise},
\end{cases}
\end{equation}
where $r_r>0$ rewards remission, $r_a=-r_r$ penalizes adverse events, $c_a$
encodes treatment cost, and $c_o$ weights symptom burden. Episodes start from an
initial distribution derived from the stationary mixture of the disease graph and
terminate at remission, an adverse event, or the maximum treatment horizon. Each
environment seed instantiates a different disease graph, symptom distribution, and
treatment parameterization. This diversity is why the main text reports averages
across eight simulated EpiCare environments.

\begin{figure}
	\centering
\begin{minipage}{0.3\textwidth}
    \centering
    \includegraphics[width=\linewidth]{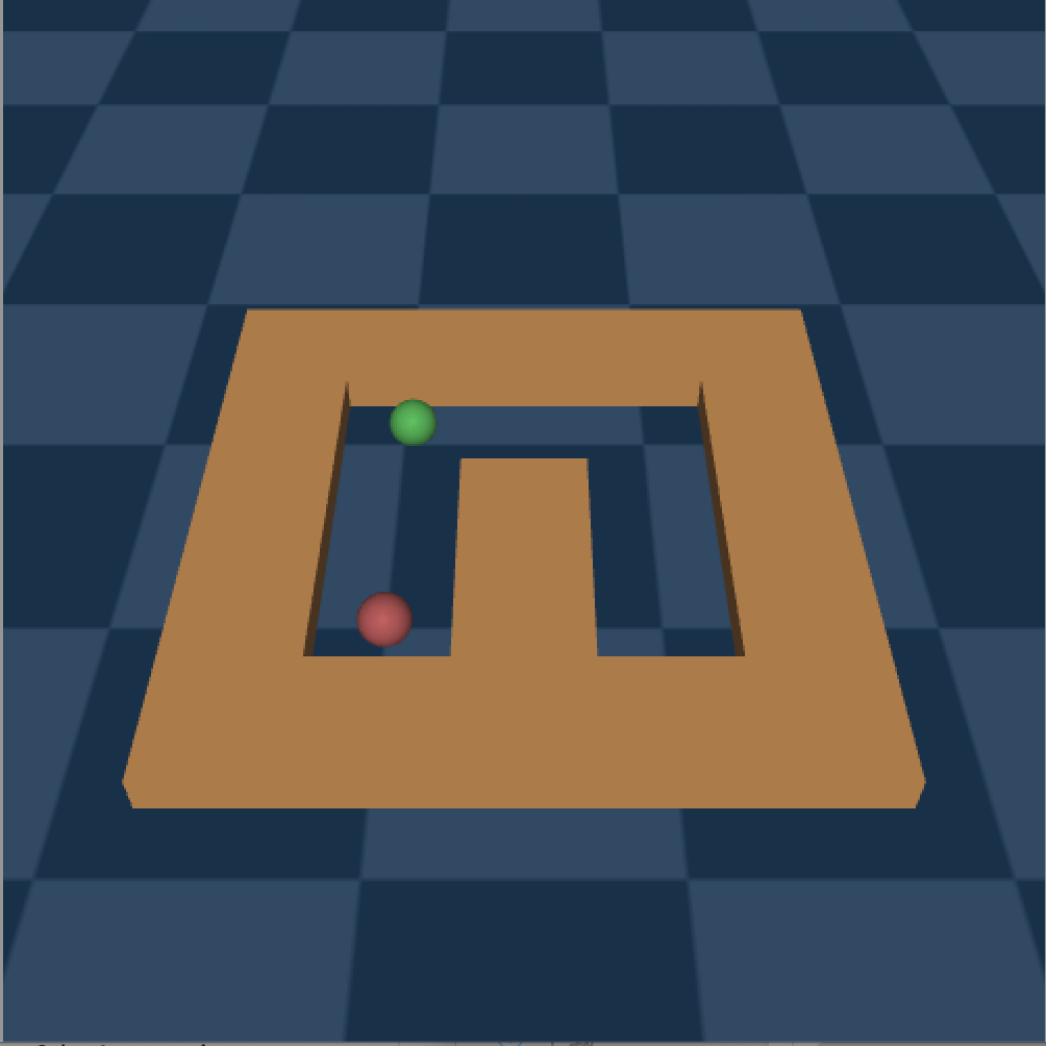}
\end{minipage}%
\begin{minipage}{0.3\textwidth}
    \centering
    \includegraphics[width=\linewidth]{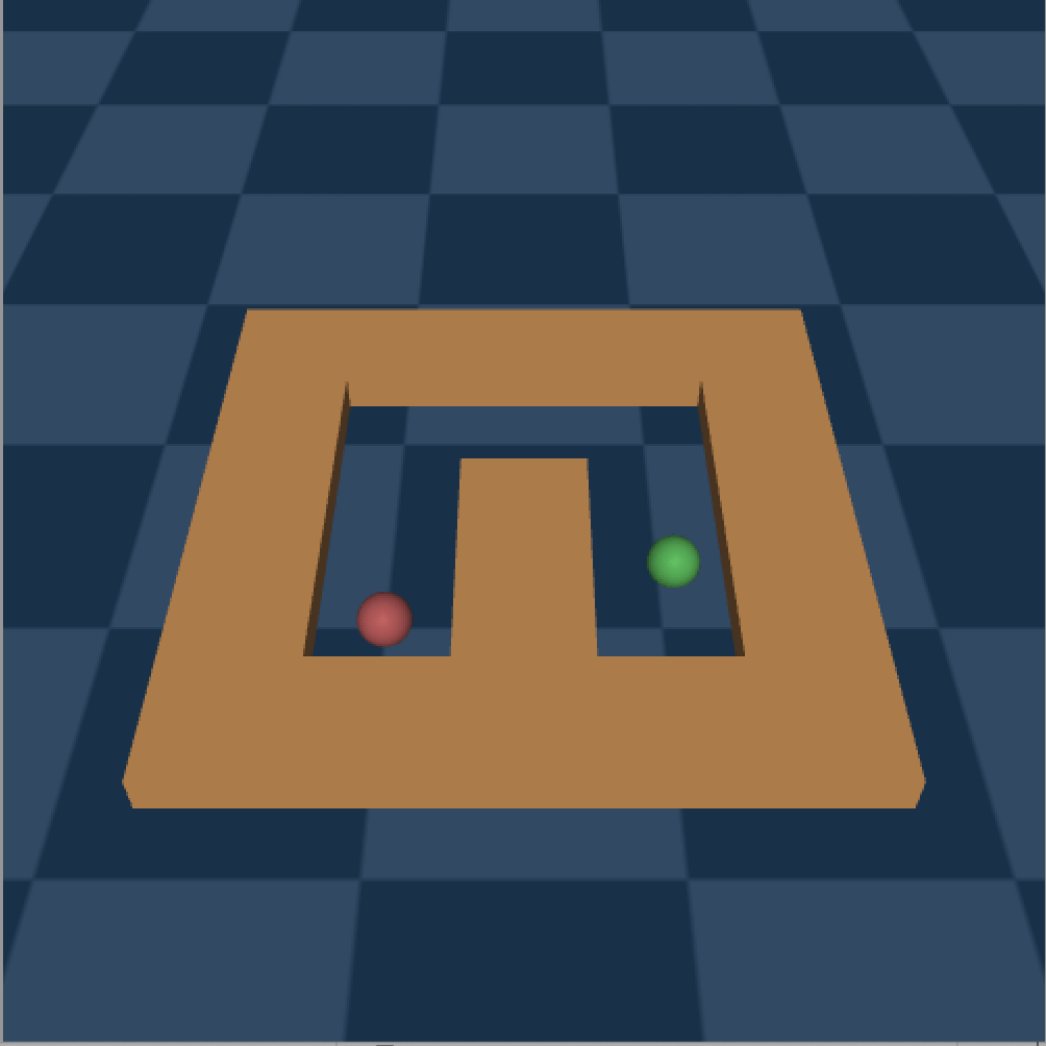}
\end{minipage}%
\begin{minipage}{0.3\textwidth}
    \centering
    \includegraphics[width=\linewidth]{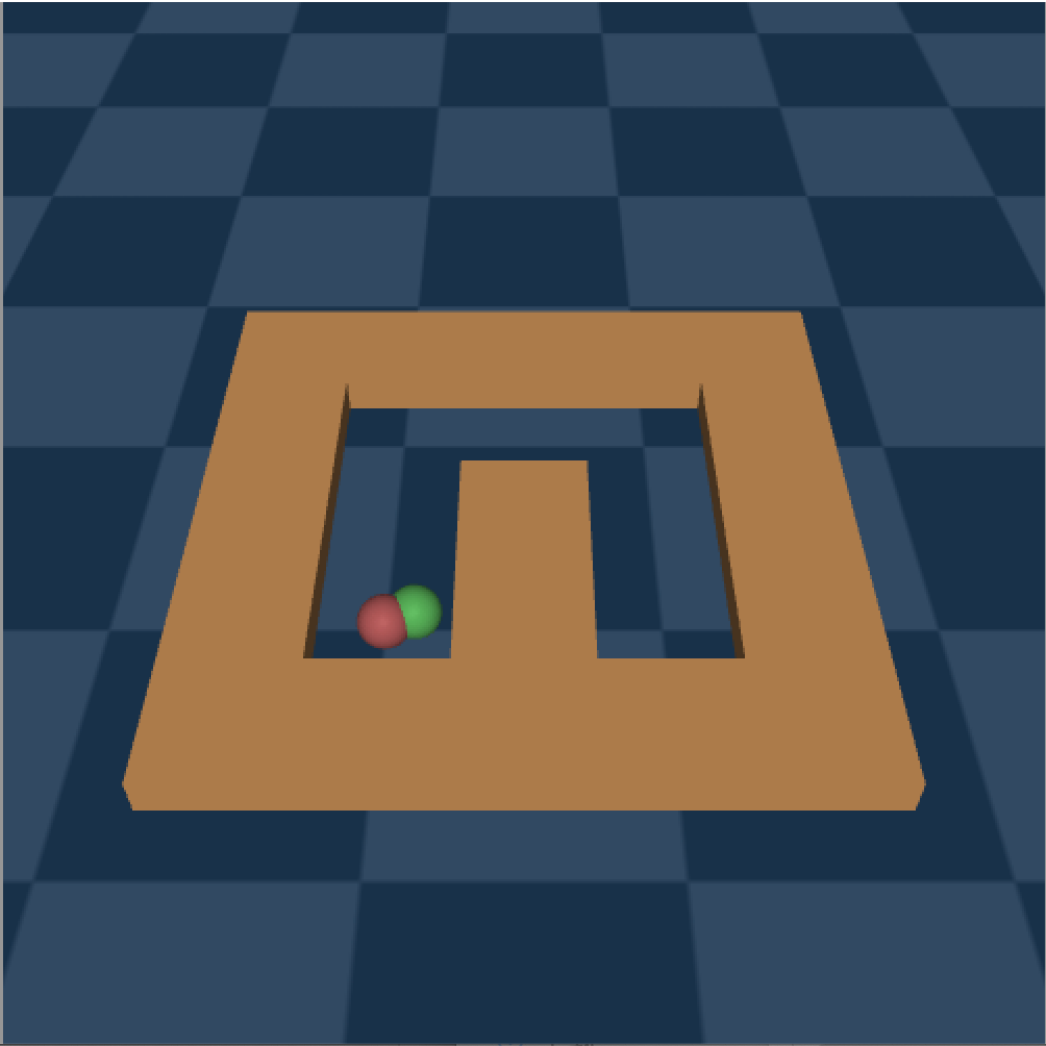}
\end{minipage}

\vspace{0.2cm}

\begin{minipage}{0.3\textwidth}
    \centering
    \includegraphics[width=\linewidth]{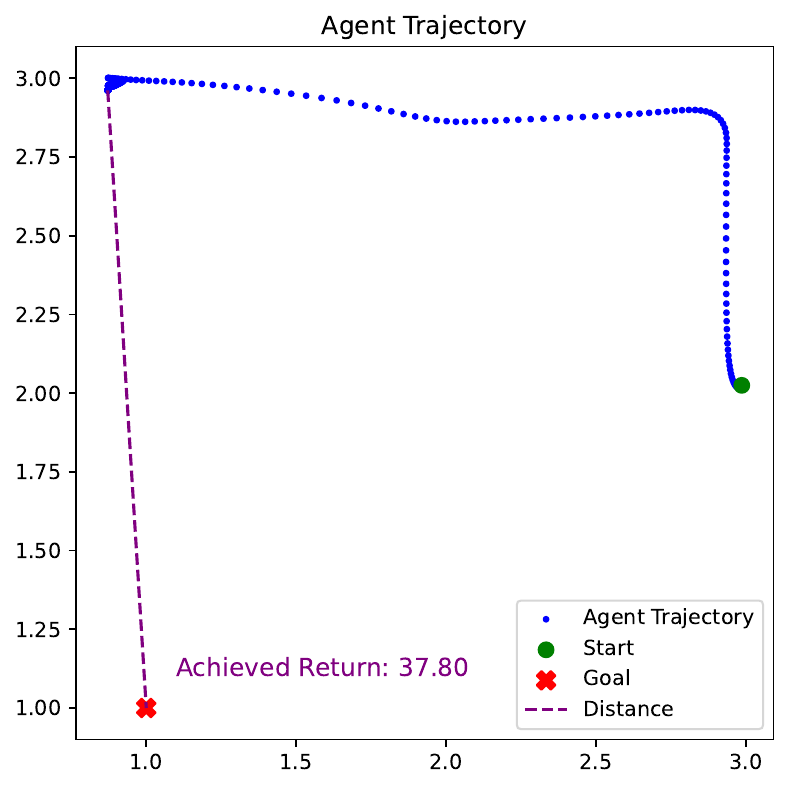}
\end{minipage}%
\begin{minipage}{0.3\textwidth}
    \centering
    \includegraphics[width=\linewidth]{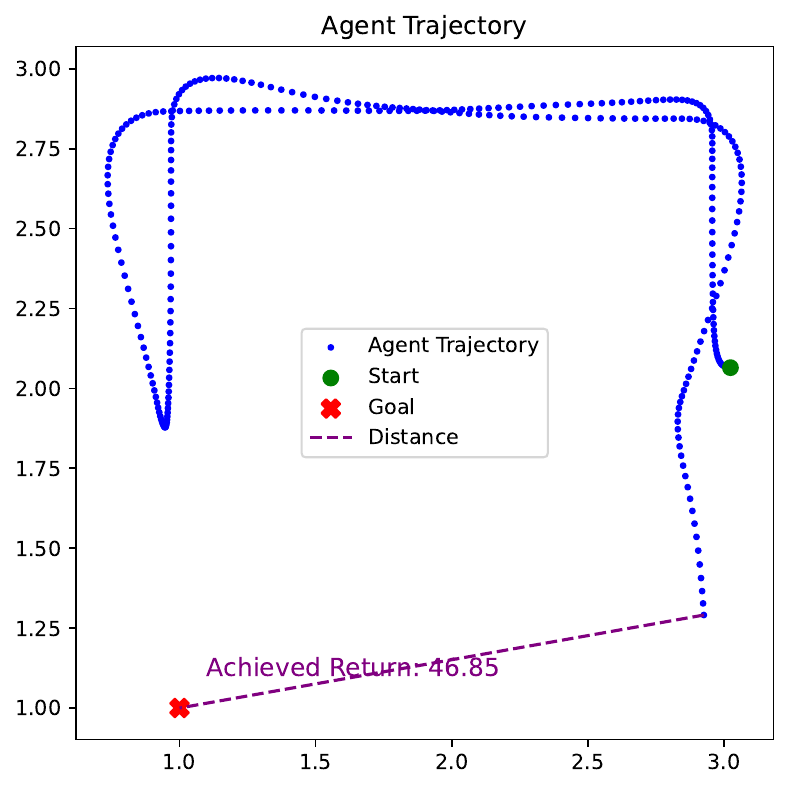}
\end{minipage}%
\begin{minipage}{0.3\textwidth}
    \centering
    \includegraphics[width=\linewidth]{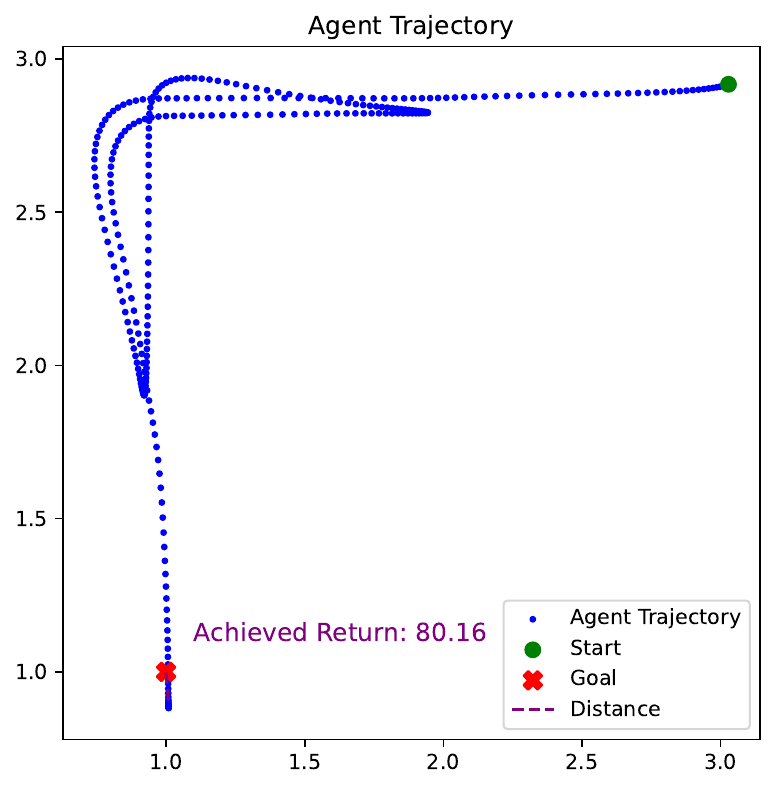}
\end{minipage}
\caption{Qualitative visualization on maze2d-umaze-dense under increasing target
returns. The first row shows frames near the end of the rollouts, and the second
row shows the corresponding trajectories. Increasing the target return moves the
agent toward higher-return behavior and closer goal-reaching trajectories.}
\label{fig:toy_example}
\end{figure}

\paragraph{behavior policy.}
We use the Standard-of-Care (SoC) policy to collect the offline dataset in the
EpiCare experiments. SoC emulates a risk-averse clinician who does not observe the
hidden disease state and instead updates treatment preferences from observed
immediate outcomes. It initializes the estimated value of treatment $a$ as
$Q_0(a)=\mathbb E[R\mid a]$ and updates the selected treatment by an exponential
recency-weighted rule,
\begin{equation}
Q_{t+1}(a)=
\begin{cases}
Q_t(a)+\alpha\left(R_t-Q_t(a)\right), & a=a_t,\\
Q_t(a), & a\ne a_t.
\end{cases}
\end{equation}
To reduce adverse events, SoC selects greedily within a safe action set,
\begin{equation}
A_{\mathrm{safe}}(o)=
\left\{a\in A:(\delta_a)_i\le0\ \text{for all}\ i\ \text{with}\
o_i\ge1-\frac{\kappa}{2}\right\}.
\end{equation}
Thus, the behavior policy avoids treatments that would further increase symptoms
already close to the danger threshold $\kappa$. This construction makes EpiCare
suitable for evaluating controllable target-conditioned policies: changing the
target return should shift the learned policy between conservative and aggressive
operating points, but the benchmark remains a simulator rather than a substitute
for clinical validation.

% #####################################################################
% \section{Additional experimental results}
% \label{app:additional-results}
% \label{app:appenD}
% #####################################################################

\begin{table}[h]
\centering
\begin{tabular}{lll}
\toprule
\textbf{Transformer policy} & & \textbf{Value} \\
\midrule
Encoder layers & & 2 \\
Decoder layers & & 1 \\
Activation function & & GELU \\
Number of attention heads & & 4 \\
Embedding dimension & & 512 \\
Layers in decoding head & & 2 \\
Dropout & & 0.10 \\
Positional encoding & & Yes \\
Learning rate & & $1\times10^{-4}$ \\
Weight decay & & 0.005 \\
AdamW betas & & $[0.9,0.999]$ \\
Learning-rate warmup steps & & 20,000 \\
\midrule
\textbf{$Q$ value heads} & & \textbf{Value} \\
\midrule
Number of layers & & 2 \\
Activation function & & ReLU \\
Hidden dimension & & 256 \\
Expectile parameter $\nu$ & & 0.7 (Locomotion), 0.9 (Maze2D), \\
& & 0.8 (Adroit and EpiCare) \\
Learning rate & & $1\times10^{-4}$ \\
Weight decay & & $5\times10^{-4}$ \\
\midrule
\textbf{General} & & \textbf{Value} \\
\midrule
% Evaluation episodes (D4RL) & & 10 \\
% Evaluation rollouts (EpiCare) & & 1,000 \\
Input trajectory length & & 4 \\
Training steps & & 140,000 \\
Batch size & & 512 \\
Discount factor & & 0.99 \\
Default candidate budget $N$ & & 300 \\
Default sampling bandwidth $\delta$ & & $0.05R_{\max}$ \\
\bottomrule
\end{tabular}
\caption{Hyperparameters of \ourmethod\ used in the experiments unless otherwise
specified. The candidate budget $N$ is varied only in the dedicated ablation study.}
\label{table:parameterdoctor}
\end{table}

\subsection{Qualitative visualization of target-conditioned behavior}
\label{app:toy-visualization}

Figure~\ref{fig:toy_example} provides an additional qualitative visualization on
maze2d-umaze-dense. This example is intended to illustrate how changing the target
return changes the generated behavior; quantitative target alignment is evaluated
in Section~\ref{sec:alignment}. Maze2D contains suboptimal trajectories and is
commonly used to evaluate whether offline methods can stitch trajectory fragments
into goal-reaching behavior. We evaluate an offline-trained \ourmethod\ model with
a rollout horizon of 300 steps under three target returns.

In this environment, the per-step reward is based on the exponential negative
Euclidean distance between the achieved and desired goal positions. Therefore, an
agent can receive a positive return even without reaching the goal, and different
target returns can correspond to different degrees of progress. In the
visualization, larger target return lead to trajectories that move more directly
toward the goal. This result should be read as a qualitative illustration of
controllable behavior, not as a substitute for the alignment curves in the main text.

% #####################################################################
\section{Model and training details}
\label{app:model-training-details}
\label{app:appenE}
% #####################################################################

We implement \ourmethod\ as a transformer-based policy with shared representations
for action generation and value verification. The supervised trajectory model uses
a bidirectional transformer encoder and a bidirectional transformer decoder. Each
input modality is projected into the embedding space by an independent embedding
layer. The decoder output is connected to a two-layer MLP with layer normalization
for trajectory reconstruction.

The transformer is trained with randomly sampled mask ratios following masked
trajectory modeling~\citep{wu2023masked}. We use
\begin{equation}
\mathrm{mask\_ratios}=[0.60,0.70,0.80,0.85,0.90,0.95,1.00].
\end{equation}
For data sampling, we follow the two-step strategy used in Decision Transformer:
we first sample a trajectory from the offline dataset and then uniformly sample a
length-$K$ sub-trajectory from it~\citep{chen2021decision}. Unless otherwise
specified, $K=4$.

For offline training, we perform 140,000 gradient updates and report final D4RL
results over five random seeds. We evaluate D4RL policies by rolling out 10
episodes per evaluation. For EpiCare, we use 1,000 rollouts because the benchmark
reports both clinical return and adverse-event rates. At inference, we use
$N=300$ candidate prompts and set the sampling bandwidth to
$\delta=0.05R_{\max}$, where $R_{\max}$ is the maximum return in the offline
dataset. The effect of varying $N$ is reported separately in
Figure~\ref{fig:hyperparameter_N}; the coverage analysis should be interpreted as
a conditional explanation of this trend, not as an empirical proof of an exact
$O(1/N)$ rate.

We optimize the transformer parameters with AdamW and the $Q$ value head with
Adam, using warmup and decay schedules. The $Q$ value head consists of two
256-dimensional MLP layers connected to the transformer decoder representation.
For baseline implementations, we follow the settings inherited from CORL whenever
applicable~\citep{tarasov2024corl}.

\printcredits

\bibliographystyle{cas-model2-names}

\bibliography{cas-refs}



\end{document}